\documentclass[journal]{IEEEtran}

\usepackage{cite}
\usepackage{hyperref}
\usepackage{amsmath}
\usepackage{verbatim}
\usepackage{multirow}
\usepackage{array}
\usepackage{diagbox}[2011/11/22]
\usepackage{mathrsfs}
\usepackage{amsfonts}
\usepackage{mathtools}
\usepackage{multirow}
\usepackage{xcolor}
\usepackage{amssymb}
\usepackage{url}
\usepackage{makecell}
\usepackage{comment}
\ifCLASSINFOpdf
\usepackage[pdftex]{graphicx}
\else
\fi

\usepackage[textsize=tiny]{todonotes}

\newcommand{\reffig}[1]{Fig. \ref{#1}}
\newcommand{\refsec}[1]{Section \ref{#1}}
\newcommand{\reftab}[1]{Table \ref{#1}}
\newcommand{\refequ}[1]{Eqn. \ref{#1}}
\newcommand{\tabincell}[2]{\begin{tabular}{@{}#1@{}}#2\end{tabular}}

\renewcommand{\tabincell}[2]{\begin{tabular}{@{}#1@{}}#2\end{tabular}}

\definecolor{red}{RGB}{255,0,0}

\begin{document}
\title{Universal Background Subtraction based on Arithmetic Distribution Neural Network}

\author{Chenqiu Zhao,  ~\IEEEmembership{Student Member,~IEEE} \thanks{The authors are with the Dept. of Computing  Science and UAHJIC, University  of Alberta, Edmonton, AB T6G 2E1, Canada (Contact e-mails: zhaochenqiu@gmail.com, zhao.chenqiu@ualberta.ca, hukang\_cmu@126.com,
 and basu@ualberta.ca)} Kangkang Hu and Anup Basu, ~\IEEEmembership{Senior Member,~IEEE} }

\maketitle

\begin{abstract}
We propose a universal background subtraction framework based on the Arithmetic Distribution Neural Network (ADNN) for learning the distributions of temporal pixels. In our ADNN model, the arithmetic distribution operations are utilized to introduce the arithmetic distribution layers, including the product distribution layer and the sum distribution layer. Furthermore, in order to improve the accuracy of the proposed approach, an improved Bayesian refinement model based on neighboring information, with a GPU implementation, is incorporated. 
In the forward pass and backpropagation of the proposed arithmetic distribution layers, histograms are considered as probability density functions rather than matrices. Thus, the proposed approach is able to utilize the probability information of the histogram and achieve promising results with a very simple architecture compared to traditional convolutional neural networks. Evaluations using standard benchmarks demonstrate the superiority of the proposed approach compared to state-of-the-art traditional and deep learning methods.
To the best of our knowledge, this is the first method to propose network layers based on arithmetic distribution operations for learning distributions during background subtraction.
\end{abstract}


\IEEEpeerreviewmaketitle

\section{Introduction}
\label{sec_introduction}

Background subtraction is a fundamental research topic in computer vision, which has attracted increasing attention during a period of explosive growth in video streaming.
Recently, several sophisticated models based on deep learning networks have achieved excellent performance.
Unfortunately, to the best of our knowledge, there are still a few challenges that limit the use of deep learning networks in real applications of background subtraction.
First, such algorithms usually require a large number of ground-truth frames for training; however, creating ground-truth frames is quite expensive since every pixel of each frame has to be labelled.
Moreover, several excellent networks for background subtraction perform poorly for unseen videos, because of their dependence on the scene information in training videos.
Finally, various networks are trained with different videos and the parameters of networks for different testing videos are thus different.
Therefore, there is no single well-trained network that can be applied for all testing videos.
In order to address these challenges, a universal background subtraction method based on the Arithmetic Distribution Neural Network (ADNN) is proposed.
\begin{figure}[!t]
    \centering
    \includegraphics[width=0.99\linewidth]{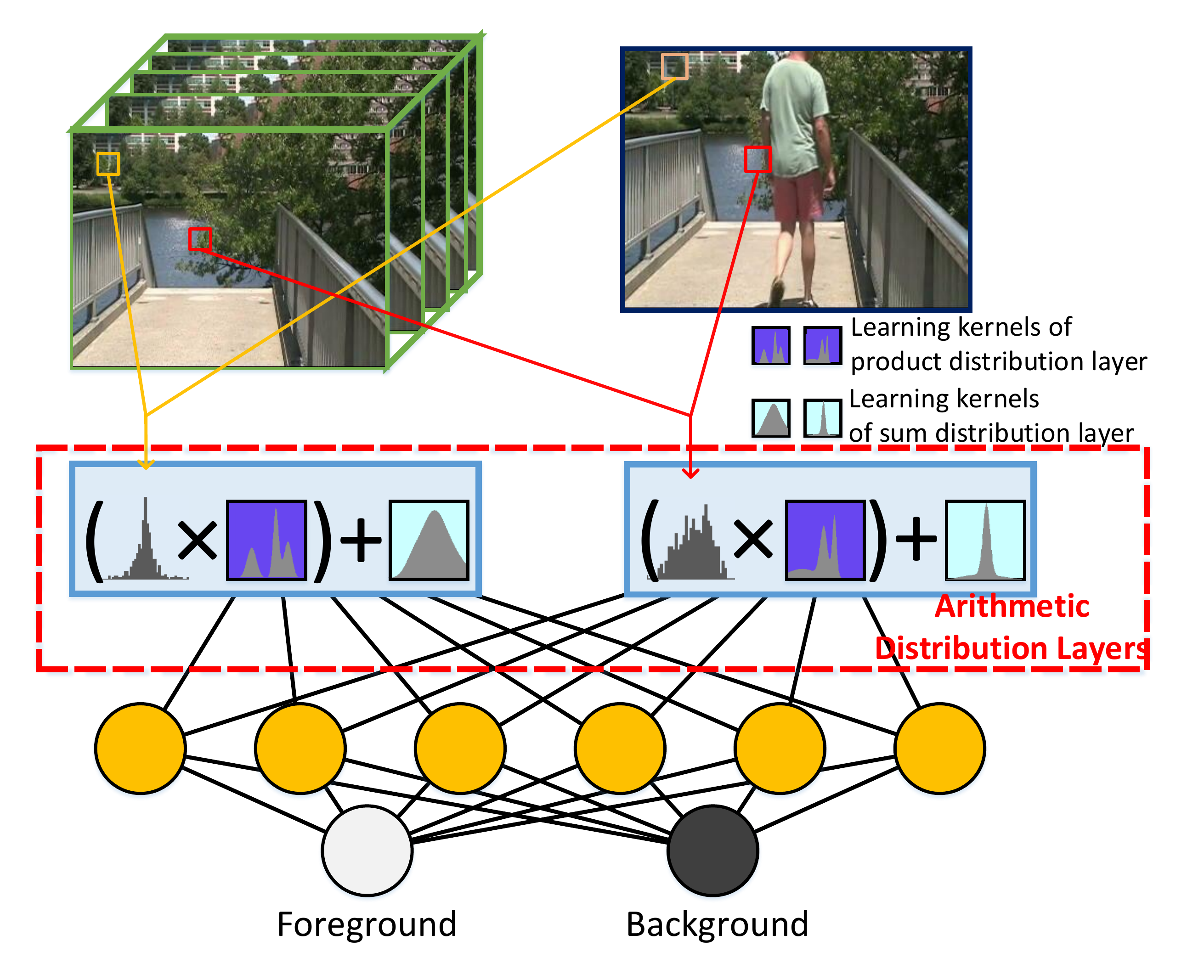}
    \DeclareGraphicsExtensions.
    \vspace{-20pt}
    \caption{Illustration of the arithmetic distribution neural network for background subtraction. 
    Histograms of subtractions between the current observations and their historical counterparts in pixels are input into the arithmetic distribution layers, containing the product and sum distribution layers for distribution learning.
    In particular, the learning kernels of arithmetic distributions layers are also distributions described by histograms.
    A classification architecture is then attached to label the pixels according to the output of these layers.}
    \label{fig1}		
\end{figure}

In background subtraction, pixels are classified as foreground or background based on comparisons with their historical counterparts.
Thus, previous approaches captured a background image to represent the historical observations of pixels.
Several recent methods used deep learning networks to learn the background representation for subtraction.
However, such networks usually require a large number of ground-truth frames to learn the background representation and need to train a particular network for every video, to deal with the diversity in scene information for different videos.
In addition, although a few excellent research (e.g., FgSN \cite{2018_PRL_LIM2018256}) have achieved almost perfect results, their performance declines when an unseen video is introduced for segmentation.
Background subtraction is essentially a classification of temporal pixels. The distributions of comparisons between temporal pixels are useful features that can be directly input into the network for classification.
Therefore, in this paper, we focus on learning the distribution of comparisons to classify pixels into foreground or background,
based on a new Arithmetic Distribution Neural Network (ADNN), as shown in \reffig{fig1}.

The architecture of the proposed network is straightforward.
Distributions described by histograms of subtractions between pixels' current observations and their historical counterparts are used as the input of the arithmetic distribution neural network.
In particular, we propose the arithmetic distribution layers including the product and sum distribution layers as the first block of the network for learning distributions.
A classification block is attached to label pixels according to the output of the arithmetic distribution layers.
The architecture of the classification block is kept as simple as possible, with only one convolutional layer, one rectified linear unit layer, and one fully connected layer, to demonstrate that the good performance of the proposed approach comes from the proposed arithmetic distribution layers.
Unfortunately, since pixels are classified independently,
the proposed network is sensitive to noisy points that can be handled by neighboring information. An improved Bayesian refinement model, with a GPU implementation, is thus proposed for noise compensation.
By utilizing the arithmetic distribution layers, histograms are considered as probability density functions, with the probability information being utilized.
This helps the proposed approach achieve better distribution learning ability compared to even convolutional neural networks. Furthermore, since the histograms of temporal pixels are pixel-wise features, a large number of training instances can be captured. Thus, the proposed approach requires fewer than 1\% of the ground truth frames during training.
Finally, since the distribution information of temporal pixels is independent of scene information, the proposed network does not rely too much on the scenes where training frames are captured. 
Our ADNN can be trained with video frames obtained from different scenes, and it is valid even when no frame from the scenes of the testing videos is included in the training set.
In addition, the independence of distribution information allows us to train only one network for all seen and unseen videos.
The main contributions of this paper are:
\begin{itemize}
    \item We propose the Arithmetic Distribution Neural Network (ADNN) for background subtraction, utilizing the product distribution layer and the sum distribution layer.
    \item An improved Bayesian refinement model, with a GPU implementation, is proposed to improve the accuracy of our approach. In particular, an approximation of the Gaussian function is utilized to compute the correlation between neighboring pixels.
    \item Comprehensive experiments are conducted to evaluate the proposed approach, including:
        a) comparisons between the proposed ADNN and traditional convolutional neural networks on real data, as shown in \refsec{sec_comp_CNN};
        b) an ablation study of the proposed arithmetic distribution layers;
        c) a comprehensive comparison between the proposed approach and state-of-the-art methods including traditional and deep learning approaches on standard benchmarks, as shown in \refsec{sec_evaluation}.
\end{itemize}

\section{Related Work} 
In this section, we outline background subtraction algorithms briefly into two categories.
In \refsec{rel_tra}, the algorithms without using a deep learning network are introduced.
Then, several excellent deep learning networks for background subtraction are discussed in \refsec{rel_dl}.

\subsection{Non-Deep-Learning Algorithms}
\label{rel_tra}
Background subtraction is a fundamental research topic in computer vision\cite{vaswani2018robust,narayanamurthy2018fast,bouwmans2021emerging,minematsu2018analytics},
and a large number of methods have been proposed on this topic \cite{Barnich2011_2011_TIP, 2005_TPAMI_1407887,
Zivkovic2004, 2004_TIP_1344037, 2009_TM_4796296,
2014_TPAMI_6678500, 2014_CVPRW_6910016, 2016_NC_RAMIREZALONSO2016990,
LIANG20151374, 2017_ICIAP_combing, 2015_ICME_7177419,
2015_TIP_6975239, 2017_TCSVT_7911235, 2018_TPAMI_7994669, 2019_TIP_8625603,bouwmans2010statistical, goyal2018review,
Stauffer1999, Kim2004, 2011_JACM_RPCA,
2017_TIP_8017547, 2019_TIP_8485415,2020_TIP_9066884,
2017_TCSVT_7434630,2017_TCSVT_7938679,GARCIAGARCIA2020100204, 2012_ECCV_Haines2012, 2018_TPAMI_7954680, 2014_ECCV_Chen2014, 2018_IS_AKILAN2018414, 2010_ICME_5582598,2010_CVPR_Liao2010,2016_TIP_7557059,2017_TFS_7468482,2015_PR_Varadarajan20153488,Elgammal2000, 2012_TIP_6203582, 2012_ICSP_6491583}.
Throughout the development of background subtraction algorithms,
the distribution of temporal pixels has played an important role since it is a good representation of background information.
In particular, the Gaussian mixture model proposed by Zivkovic et al. \cite{Zivkovic2004} is one of the most popular techniques,
where the background observations are described by several Gaussian functions \cite{Zivkovic2004, Stauffer1999}, with a large number of extensions being proposed \cite{bouwmans2010statistical, goyal2018review}.
For example, Lee et al. \cite{2005_TPAMI_1407887} utilized an adaptive learning rate for each Gaussian function to improve the convergence rate during clustering.
Haines et al. \cite{2014_TPAMI_6678500} \cite{2012_ECCV_Haines2012} used the Dirichlet processes with Gaussian mixture models to analyze pixel distributions.
Recently, Chen et al. \cite{2018_TPAMI_7954680} \cite{2014_ECCV_Chen2014} used Gaussian mixture models to represent the vertices of spanning trees and Akilan et al. \cite{2018_IS_AKILAN2018414} proposed a foreground validation process through probability estimation of multivariate Gaussian model distribution. Besides the Gaussian distribution, there are also several other techniques for the description of temporal pixels,
such as Laplacian distribution \cite{2010_ICME_5582598}, kernel density estimation \cite{2010_CVPR_Liao2010} and artificial neural networks \cite{2016_TIP_7557059}.

In addition, several excellent publications \cite{2017_TIP_8017547,2019_TIP_8485415,2011_JACM_RPCA,2018_TPAMI_7994669} have considered the background as a low-rank component of video frames, given the correlation between background scenes of frames over time.
For example, Javed et al. \cite{2017_TIP_8017547, 2019_TIP_8485415} utilized robust principal component analysis \cite{2011_JACM_RPCA} to separate the background scenes based on the spatial and temporal subspaces. Yong et al. \cite{2018_TPAMI_7994669} proposed online matrix factorization for background subtraction.
Machine-learning techniques have also been utilized for background subtraction
\cite{2002_ICIP_1039116, 2009_ICCV_5459454, 2012_TPAMI_Han2012, 2014_ICME_6890245,2017_TM_7920340,2007_TNN_4359175,2013_TPAMI_6216381,2011_ICME_6012085, 6910014,2008_TIP_4527178, 2012_CVPRW_6238922}.
Lin et al. \cite{2002_ICIP_1039116} classified the pixels using a probabilistic support vector machine.
Similarly, Han et al. \cite{2012_TPAMI_Han2012} used density-based features into a classifier utilizing a support vector machine.
Li et al. \cite{2009_ICCV_5459454} formulated background subtraction as minimizing a constrained risk function
and Culibrk et al. \cite{2007_TNN_4359175} proposed an unsupervised Bayesian classifier using a neural network architecture for background subtraction.
Unfortunately, given the complexity and diversity of natural scenes, artificial models are not adequate to generate promising classifications of pixels for all videos captured from natural scenes.
Thus, recent research has focussed on using deep learning networks to perform background subtraction automatically.

\subsection{Algorithms based on Deep Learning} 
\label{rel_dl}
Since convolutional neural networks have demonstrated an excellent ability to learn scene information,
several approaches \cite{2014_CDN_6910011,2015_TIP_6975239,2016_IWSSIP_7502717,2016_PRL_Wang,2018_ICME_8486510,2018_ICME_8486556,2018_IEEE_8319408,2018_PRL_LIM2018256,2018_PR_BABAEE2018635,2018_arXiv_lim,2019_ACCESS_8768285,2019_JEI_bgconv,2019_NN_BOUWMANS20198,2019_TCSVT_8892609,2019_WACV_mondejar2019end,2020_FCV_19,2020_ICIP_9191151,2020_ICIP_9191235,2020_NC_ZHENG2020178,2020_TCSVT_9281081,2020_TIP_9263106,2020_TITS_9238403,2020_TPAMI_9288631,2020_WACV_Tezcan,2021_ACCEE_9395443,6910014,CUEVAS2016103,2021_Neurocomputing_HUANG202184,He_2017_ICCV,Patil_2021_WACV,deng2009imagenet,krizhevsky2012imagenet,ul2018vgg16,2021_TIP_9529020}
have used deep learning networks to learn the background scenes for subtraction. For example, Wang et al. \cite{2016_PRL_Wang} proposed a fully connected network to learn the background scenes.
Zeng et al. \cite{2018_IEEE_8319408} utilized a multi-scale strategy to improve the results. 
Similarly, Lim et al. \cite{2018_PRL_LIM2018256} used a triplet convolutional neural network to extract multi-scale features from background scenes and Yang et al. \cite{2019_ACCESS_8768285} improved the robustness of their method by using an end-to-end multi-scale Spatio-temporal (MS-ST) method to extract deep features from scenes.
Unfortunately, these papers usually assume a large number of ground truth frames for training, which is very expensive in background subtraction applications.
In contrast, Babaee et al. \cite{2018_PR_BABAEE2018635} proposed a robust model in which a network is used to subtract the background from the current frame, using only 5\% of the labeled masks for training.
Liang et al. \cite{2018_ICME_8486556} utilized the foreground mask generated by the SubSENSE algorithm \cite{2015_TIP_6975239} rather than manual labeling for training, and Zeng et al. \cite{2019_JEI_bgconv} used a convolutional neural network to combine several background subtraction algorithms together.
However, since these approaches rely substantially on the scene information, their performance decline considerably when an unseen video is tested for background subtraction.
Recently, Mandal et al. \cite{2020_TIP_9263106} incorporated temporal information by using a foreground saliency reinforcement block, and proposed the 3DCD network for unseen videos.
Similarly, Tezcan et al. \cite{2020_WACV_Tezcan,2021_ACCEE_9395443} trained a U-net architecture to do subtractions between background images and current frames,
and Giraldo \cite{2020_TPAMI_9288631} solved an optimization problem of graph signals for background subtraction, utilizing a deep learning network for feature extractions.
Such algorithms are effective in unseen videos since they use temporal information; however, a large number of ground-truth frames are still needed from other videos for training.
In this work, we focus on learning the distributions of temporal pixels for background subtraction by the proposed arithmetic distribution neural network (ADNN).
The proposed ADNN requires less than 1\% of the frames as ground-truth for training, and it is also effective for unseen videos.
In addition, given the generality of distribution information,
one ADNN can be trained for all seen or unseen videos and the parameters of networks are thus fixed for all testing videos.

Although the focus of the proposed approach is related to the D-DPDL method proposed by Zhao et al. \cite{2019_TCSVT_8892609},
there are several differences between our ADNN and the D-DPDL method proposed in \cite{2019_TCSVT_8892609}.
The main difference between the proposed ADNN and D-DPDL is the mathematical method used for learning distributions.
D-DPDL devised a convolutional neural network and the entries of input patches were randomly permuted to force the network to learn distributions.
In the proposed ADNN, arithmetic distribution layers are proposed for learning distributions, which demonstrate better ability to classify distributions compared to the convolutional layers in the experiments presented in \refsec{sec_comp_CNN}.
In addition, the input of the proposed ADNN and D-DPDL are also completely different.
In D-DPDL, the input of the network is patches of temporal pixels that are randomly permuted.
By contrast, the input of the proposed ADNN is the histograms of subtractions between current pixels and their historical counterparts, since the histogram is a proper discrete representation of the probability density function which is used for computations in the proposed arithmetic distribution layers.
In addition, the number of temporal pixels is limited by the size of patches in the D-DPDL method, but the information of all temporal pixels at a particular location can be compressed into one histogram, which has better global information.
Finally, there is a big difference between the number of parameters used in D-DPDL and the proposed ADNN.
According to the network architecture published in \cite{2019_TCSVT_8892609}, there are around 7 million learning parameters in the D-DPDL model.
In contrast, as shown in \reftab{tab_net}, the ADNN devised for comparisons with state-of-the-art methods in this paper contains only around 0.1 million parameters.

\section{Arithmetic Distribution Layers}
\label{sec_adl}
In this section, the mathematical details of the proposed arithmetic distribution layers are discussed.
To the best of our knowledge, distributions have to be converted to histograms for convolutions in which the matrix arithmetic operations are used; and, all the objects involved in the operations are considered as vectors.
Under this condition, the correlation between the entries of a histogram as well as their probability information are ignored.
Essentially, histograms can be considered as the discrete approximation of the probability density functions that describe the distributions of the observed values of random variables.
When a histogram of a distribution is input into a network for classification, it can be considered as a classification of random variables that have the input distributions.
Based on this insight, we assume that the arithmetic distribution operations \cite{springer1979algebra} are better than matrix arithmetic operations for distribution analysis,
because histograms are considered as distributions rather than vectors during arithmetic distribution operations.
%
%
Thus, a new type of network layers named arithmetic distribution layers, which contain the product and sum distribution layers, is proposed as a better substitute for the convolution layers for distribution classification.
During the forward pass of the proposed arithmetic distribution layers, the input distributions are computed with the distributions in the learning kernels to generate the output distributions.
In contrast to the backpropagation of the arithmetic distribution layers, the gradient of the distributions in the learning kernels with respect to the network output is computed to update the learning kernels.
In particular, all these distributions are described by histograms and all computations are based on the arithmetic distribution operations in the proposed arithmetic distribution layers.

\textbf{Notation}: Before discussing the mathematical formulae of the forward pass and backpropagation of the proposed arithmetic distribution layers, the notation used throughout the rest of this section is first introduced.
Let $X$ and $Z$ denote the random variables following the distributions of the input and output of the proposed arithmetic distribution layers, respectively.
Let $W$ and $B$ denote the random variables following the distributions in the learning kernels of the product distribution layer and sum distribution layer, respectively.
Let $f_X(x)$, $f_W(w)$, $f_B(b)$ and $f_Z(z)$ denote the probability density functions of random variables $X$, $W$, $B$ and $Z$, respectively, where $x$, $y$, $b$ and $z$ denote the observed values of $X$, $Y$, $B$ and $Z$, respectively.
Let $\vec{x}$, $\vec{w}$, $\vec{b}$ and $\vec{z}$ denote the histograms used to describe $f_X(x)$, $f_W(w)$, $f_B(b)$ and $f_Z(z)$, respectively.
In particular, $x_n$, $w_i$, $b_k$ and $z_j$ are the entries of histograms $\vec{x}$, $\vec{w}$, $\vec{b}$ and $\vec{z}$, respectively, where $n$, $i$, $k$, $j$ are indices.
$loss$ is the final scalar output of the network using the arithmetic distribution layers.
During training, we want to minimize the value of $loss$ to approach 0 if possible.
$\delta S \equiv \frac{\partial loss}{\partial S}$ is the gradient of variable $S$ with the respect to $loss$.
$S$ can be the entries of histograms in the learning kernels, such as $w_i$ or $b_k$.

The product distribution layer is used to compute the distribution of the product of random variables $X$ and $W$ having distributions $f_X(x)$ and $f_W(w)$, respectively.
The input of the product distribution layer is a histogram describing $f_X(x)$.
Then, the histogram of the learning kernel in the layer is used to describe $f_W(w)$.
Finally, the output of the layer is a histogram of $f_Z(z)$, which is the probability density function of the random variable $Z=XW$.
In order to implement the product distribution layer,
the expressions for $f_Z(z)$ and the gradient of $f_W(w)$ must be obtained for the forward pass and the backpropagation, respectively. This is discussed in the remaining part of this section.
The forward pass of the product distribution layer is the procedure to compute $f_Z(z)$ by the product of $f_X(x)$ and $f_W(w)$.
In order to capture the expression for $f_Z(z)$, the definition of the cumulative distribution function of $Z$ is proposed first, as shown below:
\begin{equation}
    \label{eq_1}
    \begin{split}
        F_Z(z) & \overset{def}{=} \mathbb{P}(Z \leq z) = \mathbb{P}(X W \leq z) \\ 
        = & \mathbb{P}(X W \leq z, W \geq 0^{+}) + \mathbb{P}(X W \leq z, W \leq 0^{-})  \\
        = & \mathbb{P}(X \leq \frac{z}{W}, W \geq 0^{+})+\mathbb{P}(X \geq \frac{z}{W},W \leq0^{-}) \\
        \because X&W \leq z, W  \in [-\infty \ 0^{-} ] \cup [0^{+} \ \infty ]    \\
        \Rightarrow & X \leq \frac{z}{W}, W \geq 0^{+} \ or \ X \geq \frac{z}{W},W \leq0^{-}  \\
    \end{split}
\end{equation}
where $F_Z(z)$ is the cumulative distribution function of the random variable $Z$.
$\mathbb{P}$ is a cumulative distribution under a particular condition.
Next, assuming $X$, $W$ and $Z$ are between negative infinity and positive infinity, the expression of $F_Z(z)$ is converted into an expression following the cumulative distribution function.
Mathematically, this can be shown as:
\begin{equation}
    \begin{split}
        F_Z(z) & \! = \!  \int_{0^{+}}^{\infty}\! \! \! \! \! f_W(w) \! \int_{-\infty}^{\frac{z}{w}} \! \! \! \! f_X(x)dxdw  \! + \! \int_{-\infty}^{0^{-}}\! \! \! \! \! \!f_W(w) \! \int_{\frac{z}{w}}^{\infty} \! \! \! \! \! f_X(x)dxdw  \\
        \because \quad  & \mathbb{P}(X \leq \frac{z}{w}, W \geq 0^{+}) \! = \! \int_{0^{+}}^{\infty}\! \! \! \! \! f_W(w) \!  \int_{-\infty}^{\frac{z}{w}} \! \! \! \! f_X(x)dxdw       \\
        & \mathbb{P}(X \geq \frac{z}{w}, W \leq 0^{-}) \! = \! \int_{-\infty}^{0^{-}}\! \! \! \! \! \!f_W(w) \!  \int_{\frac{z}{w}}^{\infty} \! \! \! \! \! f_X(x)dxdw ,
    \end{split}
\end{equation}
where $dx$ and $dw$ are the delta of $x$ and $w$ respectively.
Then, the formula of $f_Z(z)$ can be obtained by the derivative of the cumulative distribution function $F_Z(z)$ with respect to $z$.
However, since $z$ is under the integral sign, the Leibniz integral rule is applied.
In calculus, the Leibniz integral rule is used for differentiating under the integral sign. For example, the derivative of $\int_{a(z)}^{b(z)}f(z,x)dx$ with respect to $z$,
where $-\infty < a(z), b(z) < \infty$, can be expressed as:
\begin{equation}
        \begin{split}
        \frac{d}{dz}\left( \int_{a(z)}^{b(z)}f(z,x)dx \right) & =  f(z,b(z))\cdot \frac{d}{dz}b(z) \\
         -f(z,a(z))\cdot & \frac{d}{dz}a(z)   +\int_{a(z)}^{b(z)}\frac{\partial}{\partial z}f(z,x)dx.
         \end{split}
\end{equation}
Thus, with the help of the Leibniz integral rule, the expression for $f_Z(z)$ can be obtained as the derivative of the cumulative distribution function $F_Z(z)$ with respect to $z$.
This is expressed as:
\begin{equation}
    \begin{split}
        f_Z(z) & = \frac{d(F_Z(z))}{ dz} \\
        = & d \! \!  \left( \! \int_{0^{+}}^{\infty}\! \! \! \! \! \! \! f_W(w) \! \! \int_{-\infty}^{\frac{z}{w}} \! \! \! \! \! \! f_X(x)dxdw \! + \! \! \int_{-\infty}^{0^{-}}\! \! \! \! \! \! \! \! f_W(w) \! \! \int_{\frac{z}{w}}^{\infty} \! \! \! \! \! \! \! f_X(x)dxdw \! \right) \! \! / \!dz \\
        = & \int_{0^{+}}^{\infty}\! \! \! \! \! \! \! f_W(w) \! \! \left[ \! \frac{ d \! \! \int_{-\infty}^{\frac{z}{w}} \!  f_X(x)dx }{dz} \! \right] \! \! \! dw \! + \! \! \! \! \int_{-\infty}^{0^{-}}\! \! \! \! \! \! \! \!  f_W(w) \! \! \!\left[ \! \frac{  d \! \! \int_{\frac{z}{w}}^{\infty} \! \!   f_X(x)dx}{dz} \! \right] \! \! \! dw  \\
        = & \int_{0^{+}}^{\infty}f_W(w)f_X(\frac{z}{w})\frac{1}{w}dw + \int_{-\infty}^{0^{-}}f_W(w)f_X(\frac{z}{w})\frac{1}{-w}dw \\
        = & \int_{0^{+}}^{\infty}f_W(w)f_X(\frac{z}{w})\frac{1}{|w|}dw + \int_{-\infty}^{0^{-}}f_W(w)f_X(\frac{z}{w})\frac{1}{|w|}dw \\
        = & \int_{ -\infty }^{\infty}f_W(w)f_X(\frac{z}{w})\frac{1}{|w|}dw \\
        \because &  \frac{ d  \int_{-\infty}^{\frac{z}{W}}  f_X(x)dx }{dz}=
        f_X(\frac{z}{w})\frac{d}{dz}(\frac{z}{w}) - f(-\infty)\frac{d}{dz}(-\infty) \\
        & \ \ \ \ \ \ \ +  \int_{-\infty}^{\frac{z}{W}}\frac{\partial}{\partial z}(  f_X(x)dx) = f_X(\frac{z}{w})\frac{1}{w} - 0 + 0 \\
        &   \frac{  d  \int_{\frac{z}{W}}^{\infty}   f_X(x)dx}{dz}=
        f_X(\infty)\frac{d}{dz}(\infty)-f_X(\frac{z}{w})\frac{d}{dz}(\frac{z}{w}) \\
        & \ \ \ \ \ \ \ +  \int_{\frac{z}{W}}^{\infty}\frac{\partial}{\partial z}(f_X(x)dx) = 0 - f_X(\frac{z}{w})\frac{1}{w} + 0 \\
        & |w| = -w \ if \ w \in [-\infty \ 0^{-}], \ |w| = w, \ if \ w \in [0^{+} \ \infty] \\
        \Rightarrow z_j & = \sum\limits_{i=-\infty}^{\infty}w_i f_X(\frac{z_j}{i})\frac{1}{|i|} \cdot 1 \quad \because dw = 1,\  f_W(i) = w_i, 
    \end{split}
    \label{eq_product_distribution}
\end{equation}
where $w_i$ and $z_j$ are the entries of histograms $\vec{w}$ and $\vec{z}$ that are used to describe $f_W(w)$ and $f_Z(z)$, respectively.
The formula for the forward pass of the product distribution layer is thus derived.
Then, the gradient of $f_W(w)$, which is used to update $w_i$ during backpropagation, is obtained by partial derivatives and the chain rule.
Mathematically:
\begin{equation}
    \begin{split}
        \frac{\partial loss}{\partial w_i} & = \frac{\partial loss}{\partial Z} \cdot \frac{\partial Z}{\partial w_i} = \sum\limits_{j=-\infty}^{\infty} \frac{\partial loss}{\partial z_j} \cdot \frac{\partial z_j}{\partial w_i} \\
        & = \sum\limits_{j=-\infty}^{\infty} \frac{\partial loss}{\partial z_j} \cdot \frac{\partial \left( \sum\limits_{i=-\infty}^{\infty}w_i f_X(\frac{z_j}{i} )\frac{1}{|i|} \right)}{\partial w_i} \\
        & = \sum\limits_{j=-\infty}^{\infty} \delta z_j f_X(\frac{z_j}{i})\frac{1}{|i|} \\
        \Rightarrow \delta w_i & = \sum\limits_{j=-\infty}^{\infty} \delta z_j f_X(\frac{z_j}{i})\frac{1}{|i|} , \\
    \end{split}
    \label{eq_prodis_back}.
\end{equation}
where $\delta w_i$ and $\delta z_j$ are the gradients of entries of histograms of $f_W(w)$ and $f_Z(z)$ respectively, $i$ and $j$ are indices.
$loss$ is the final scalar output of the network having product distribution layers, which is also the output of loss functions such as Mean Squared Error or Cross-entropy Loss during training.
This way, the formula for backpropagation of the proposed product distribution layer is derived.

Similarly, the sum distribution layer is used to compute the distribution of the sum of random variables $X$ and $B$, which are described by $f_X(x)$ and $f_B(b)$, respectively. Similar to the product distribution layer, $f_X(x)$ and $f_B(b)$ are represented by histograms as well. Utilizing the same mathematical procedure as the product distribution layer, the expression of the probability density function of the sum $Z=X+B$ of $X$ and $B$ is obtained. Mathematically:
\begin{equation}
    \begin{split}
        f_Z(z) & = \int_{- \infty}^{\infty}f_B(b)f_X(z - b)db \\
        \Rightarrow z_j & = \sum\limits_{i=-\infty}^{\infty}b_i f_X(z_j - i)\cdot1 \quad \because db = 1, \  f_B(i) = b_i ,
    \end{split}
    \label{eq_addition_distribution}
\end{equation}
where $b_i$ and $z_j$ are the entries of histograms utilized to describe $f_B(b)$ and $f_Z(z)$ corresponding to random variables $B$ and $Z$, respectively. $i$ and $j$ are indices.
Also, the formula for backpropagation is:
\begin{equation}
    \begin{split}
        \frac{\partial loss}{\partial b_k} & = \frac{\partial loss}{\partial Z}\cdot \frac{\partial Z}{\partial b_k} = \sum\limits_{j = -\infty}^{\infty} \frac{\partial loss}{z_j}\cdot \frac{\partial z_j}{\partial b_k} \\
        & =  \sum\limits_{j = -\infty}^{\infty} \frac{\partial loss}{z_j}\cdot \frac{\partial (\sum\limits_{i=-\infty}^{\infty}b_i f_X(z_j - i))}{\partial b_k} \\
        & = \sum\limits_{j=-\infty}^{\infty} \delta z_j  f_X(z_j - k) \\
        \Rightarrow \delta b_k &=  \sum\limits_{j=-\infty}^{\infty} \delta z_j f_X(z_j - k), \\
    \end{split}
    \label{eq_adddis_back}
\end{equation}
where $\delta b_k$ and $\delta z_j$ are the gradients of entries in histograms corresponding to $f_B(b)$ and $f_Z(z)$ respectively, and $k$ and $j$ are indices.
$loss$ is the final output of the network using the sum distribution layer.

With the help of Eqn. \ref{eq_product_distribution}\!\! -- \!\!\ref{eq_adddis_back}, 
the forward pass and the backpropagation of arithmetic distribution layers can be easily implemented in Pytorch \cite{paszke2019pytorch}.
In particular, the gradient of the learning kernels of the arithmetic distribution layers is computed and input into the ``Autograd package'' of PyTorch \cite{paszke2019pytorch} for backpropagation.
Because of the limit on the length of this paper, the validation experiment for the proposed arithmetic distribution layers are outlined in the appendices in the supplementary material.
In our implementation of the arithmetic distribution layers,
since the distributions of temporal pixels whose values have been normalized within $[0, 1]$, are used as the network input,
the $x$ coordinate of histograms is narrowed into $[-1 ,1]$ with a bin interval of $0.01$.
This means that there are $(1 - (-1))/0.01 + 1 = 201$ bins in the histograms. This is the reason why the size of the arithmetic distribution layers shown in \reftab{tab_net} is $3 \times 201 \times 1$, where $3$ represents the RGB channels of images.
Furthermore, the sum of probability values falling into bins are normalized to 1, before the histograms are used as network input.
It should be noted that a larger range of the $x$ coordinate of histograms can give us more accurate results. However, this implementation setting is enough to generate promising results, and is used for all experiments proposed in this paper.
\begin{table}[!t]			
    \caption{Details of the proposed arithmetic distribution neural network architecture for background subtraction.}
\vspace{-5pt}
\label{tab_net}				
\centering
 \begin{tabular}{|l|l|l|l|}
\hline
     Type                   & Filters   & Layer size   & Data size     \\
\hline
     Input                  &           &                                   &   B$ \times 3 \times 201\times 1$   \\
     Product Distribution   & 2         & $ 3 \times 201 \times 1  $            &   B$ \times 3 \times201\times 2$  \\
     sum distribution       & 2         & $ 3 \times 201 \times 1 $            &   B$ \times 3   \times 201\times2$ \\
     Convolution            & 1        & $3 \times 1 \times 2 $                 &   B$ \times 10\times201 \times 1$ \\
     Convolution            & 512       &   $1\times201 \times 1$                   &   B$\times 512\times1 \times 1$  \\
     Rectified linear unit  &           &                                   &      \\
     Convolution            & 2         &   $512\times1 \times 1$                    &   B$\times 2 \times 1 \times 1$ \\
     \text{Softmax}         &           &                                   &                           \\
\hline
\end{tabular}
\raggedleft
\ \    B: Batch size. \ \
    \vspace{-10pt}
\end{table}

\section{Arithmetic Distribution Neural Network for Background Subtraction}
\begin{figure*}[!t]
    \centering
    \includegraphics[width=0.99\linewidth]{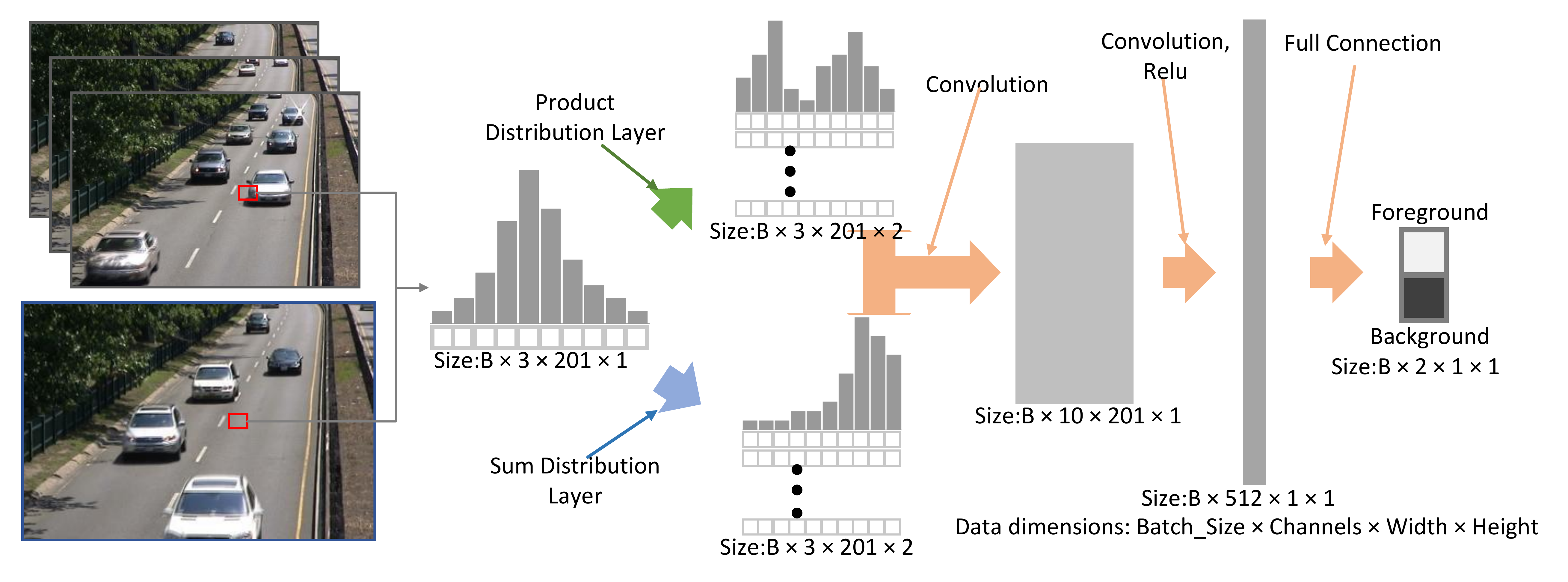}
    \DeclareGraphicsExtensions.
    \vspace{-25pt}
    \caption{An illustration of the arithmetic distribution neural network for background subtraction. Histograms of subtractions between a pixel's current observation and its historical counterparts are used as the input to arithmetic distribution layers for learning distributions.
    The output histograms of arithmetic distribution layers are combined by a convolution and input into a classification architecture containing a convolution layer, a rectified linear unit (Relu) layer, and a fully connected layer for classification.}
    \label{fig_flowchart}		
    \vspace{-10pt}
\end{figure*}
Utilizing the proposed product and sum distribution layers, the arithmetic distribution neural network is devised for background subtraction. Background subtraction is a binary classification of temporal pixels; thus, the distributions of temporal pixels play an important role.
In this work, the distributions of subtractions between pixels and their historical counterparts are used for classification.
In particular, histograms are utilized to describe the distributions of subtractions and also directly used as the input of the proposed arithmetic distribution neural network. The network architecture is quite straightforward: histograms are first input into the product distribution layer and the sum distribution layer. Then, the outputs of these layers are combined by a convolution followed by a classification architecture which consists of a convolution, a rectified linear unit (Relu) layer, and a fully connected layer. The classification architecture is deliberately kept as simple as possible, with only 3 layers, in order to demonstrate that the good results come from the proposed arithmetic distribution layers.

The components of the proposed arithmetic distribution neural network for background subtraction are illustrated in \reffig{fig_flowchart}, 
with details of the network architecture presented in \reftab{tab_net}, in which the first two convolutions are 1D convolution, since one of the values of the kernel size is 1.
Starting with a given frame of a video, denoted as $\mathcal{I} = \{I_1, I_2, \cdots, I_T\} = \{I_t | t = [1,\ T] \cap \mathbb{N}\}$, where $t$ is the frame index, $T$ is the number of frames, and $\mathbb{N}$ is the set of all natural numbers. To perform background subtraction for a particular pixel located at $(x,y)$ on frame $t$, the histogram of subtractions between pixels' current observation and their historical counterparts is captured for classification. Mathematically:
\begin{equation}
    H_{x,y}(n) = \sum\limits_{i = 1}^{T}(I_i(x,y) -I_t(x,y))\cap n,
\end{equation}
where $H_{x,y}$ is the histogram of subtractions, and $n$ is the index of entries of the histogram.
$I_i(x,y)$ denotes historical observations of the pixel located at $(x,y)$, and $I_t(x,y)$ denotes its current observation.
The distributions of subtractions are directly used as the input to the product and sum distribution layers for distribution learning.
Then, the sum of the outputs of these two layers is used as the input of the classification architecture.
Mathematically:
\begin{equation}
    \mathcal{M}(x,y)=\mathcal{L}(\mathcal{C}(\mathcal{F}_p(H_{x,y}) + \mathcal{F}_a(H_{x,y}))) ,
\end{equation}
where $H_{x,y}$ is the input histogram; $\mathcal{F}_p$ and $\mathcal{F}_a$ denote the product distribution and the sum distribution layer; $\mathcal{C}$ is the convolution procedure; and $\mathcal{L}$ is the classification architecture consisting of a convolution, a rectified linear unit, and a fully connected layer attached with a softmax function.
In particular, the negative log likelihood loss (NLLLoss) is used to update the parameters in $\mathcal{L}$, $\mathcal{C}$, $\mathcal{F}_p$ and $\mathcal{F}_a$ during the training process.
In contrast to the testing process, the arguments of the maxima (argmax) function is attached on the last layer of the classification architecture $\mathcal{L}$ to generate the label $\mathcal{M}(x,y)$ of the histogram captured from the pixel located at $(x,y)$.
\begin{figure}[!t]
    \centering
    \includegraphics[width=0.99\linewidth]{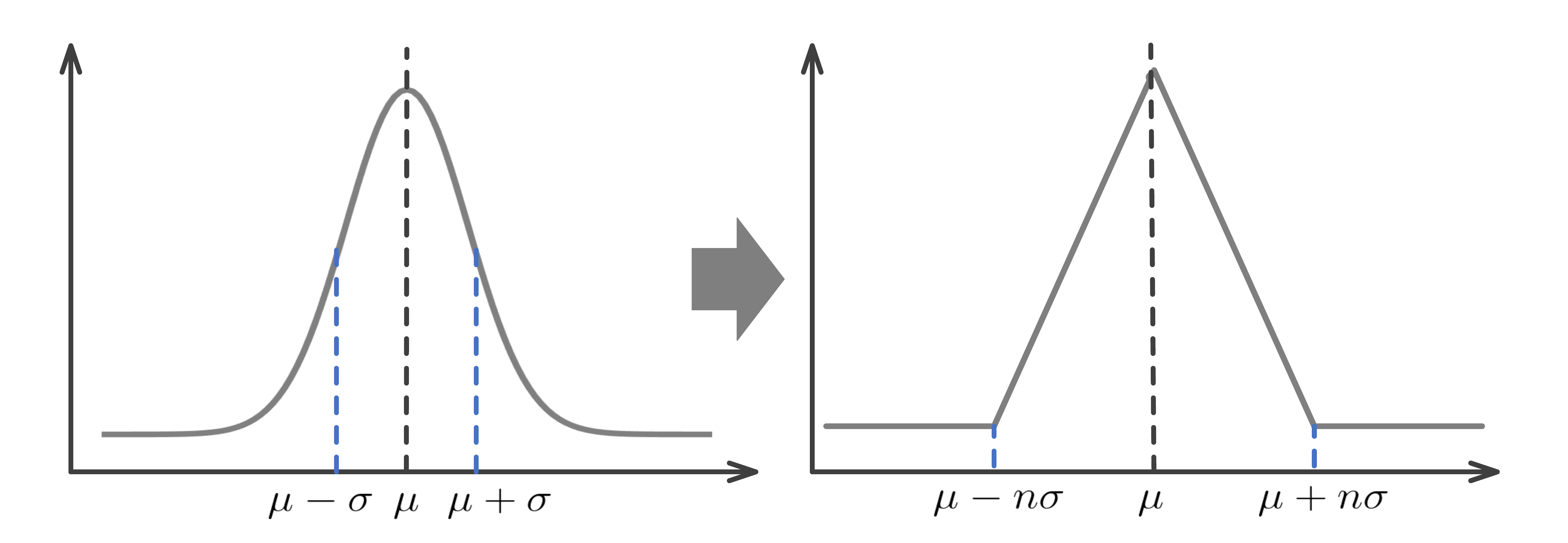}
    \DeclareGraphicsExtensions.
    \vspace{-20pt}
    \caption{An illustration of the Gaussian approximation function which approximates using a piecewise function controlled by the parameters of the Gaussian function.}
    \label{fig_approx}		
    \vspace{-10pt}
\end{figure}

Unfortunately, the histograms utilized for classification are captured from independent pixels. Thus, the correlation between pixels is ignored. In order to improve the accuracy of the proposed approach, an improved Bayesian refinement model is introduced. For completeness, we briefly introduce the Bayesian refinement model; please check \cite{2019_TCSVT_8892609} for more details. In the Bayesian refinement model, the labels of pixels are re-inferred according to the correlations with their neighborhoods, and the Bayesian theory is utilized during inference.
In particular, Euclidean distance is used to compute the correlation. In contrast, we utilize a mixture of Gaussian approximation functions to capture the correlation. This is the main difference compared to the original Bayesian refinement model. Mathematically:
\begin{equation}
    \begin{split}
        \mathcal{F}(I(x,y),&  \mathcal{M}) =  \mathop{argmax}_{a_i}\frac{P(I(x,y)|a_i)P(a_i)}{P(I(x,y))} \\
        = & \mathop{argmax}_{a_i}P(a_i)\sum\limits_{k=1}^{K}\pi_k \mathcal{N}_p (v_k|\mu_{k,i},\Sigma_{k,i}) \\ \\
        \because & P(\mathcal{I}(x,y)) = N \\ 
        & \ \ \ \& \mathop{argmax}_{x} Nx = \mathop{argmax}_{x} x \\
        & P(I(x,y)|a_i) = \sum\limits_{k=1}^{K}\pi_k \mathcal{N}_p (v_k|\mu_{k,i},\Sigma_{k,i}) ,
    \end{split}
\end{equation}
where $\mathcal{F}$ denotes the proposed improved Bayesian refinement model.
$a_i \in \{0, 1\}$ denotes the labels of foreground or background; $I(x,y)$ is a pixel located at $(x,y)$; and,
$P(I(x,y)|a_i)$ is the probability that the label of this pixel is $a_i$, which is captured though a mixture of Gaussian approximation functions $\mathcal{N}_p (v_k|\mu_{k,i} \Sigma_{k,i})$.
In particular, $v_k$ denotes the feature vector consisting of the Lab color and spatial position of the pixel $I(x,y)$, and $k$ is the index of entries in a vector.
$u_k$ and $\Sigma_k$ denote the mean and variance of features of a pixel in a local rectangular range with center at $(x,y)$ and radius $R=4$.
$\pi_k$ is the weight to mix the Gaussian approximation functions $\mathcal{N}_p$ which is mathematically expressed as:
\begin{equation}
\mathcal{N}_p(x|\mu,\sigma) = 
    \begin{cases} 
        |1 + \frac{x - \mu }{n \sigma}| & |x - u| \leq n \sigma\\ 
        0 & otherwise 
    \end{cases}
\end{equation}
where $\mu$ and $\sigma$ denote the mean and variance,
and $n$ is a user parameter.
During experiments, $n=2$ gives us the best results.
As shown in \reffig{fig_approx}, the Gaussian approximation function is actually a rough estimate of the Gaussian function.
We use a piecewise function to approximate the waveform of the Gaussian function considering the computational cost.
Also, it is more convenient for a GPU implementation, since the original Bayesian refinement model involves Euclidean distance and adaptive weights,
which significantly accelerates the refinement procedure.
Finally, the output binary mask is used in the input again to generate better results iteratively.
The Bayesian refinement model is utilized to iteratively refine the foreground mask.
In particular, the output of the arithmetic distribution neural network is used as the initial binary mask for iterative refinement.
Mathematically:
\begin{equation}
    \mathcal{M}_n(x,y) = \mathcal{F}(I(x,y), \mathcal{M}_{n-1}) ,
\end{equation}
where $n$ is the iteration number and $\mathcal{M}_{n -1}$ is the binary mask from the last iteration.
Using a GPU implementation, with the number of iterations set to 20, which is used in all the evaluation experiments proposed in this paper, the entire refinement procedure only takes a few seconds.

The improved Bayesian refinement model (IBRM) runs much faster than the Bayesian refinement model (BRM) with almost no loss in accuracy.
A comparison between them on a few frames for videos at different resolutions is shown in \reftab{tab_BRMvsIBRM}.
In particular, the running time of BRM and IBRM with iteration numbers 1, 20 and 50 are presented, as well as the $Fm$ value of their corresponding output masks after refinement.
As shown in \reftab{tab_BRMvsIBRM}, when the number of iterations is 50, although the $Fm$ value of the output mask shows obvious improvement, the run time also increases to 52s, which is too long for real applications.
In contrast, IBRM needs only 3.5s of processing time, and the $Fm$ value of the output mask is still close to the one for BRM.
Actually, improved Bayesian refinement is devised for GPU implementation with the motivation of accelerating the refinement procedure.
Thus, the superiority of the proposed IBRM is demonstrated.
For comparisons, both BRM and IRBM are run on GPU devices, all data are moved into video memory (GPU memory) to guarantee the running environment of BRM and IBRM are the same.

\begin{table}
    \caption{Comparison between Bayesian refinement model and our improved Bayesian refinement model on frames of videos at different resolutions.}
\vspace{-5pt}
    \label{tab_BRMvsIBRM}
    \centering
\begin{tabular}{|@{  }l@{  }|@{  }l@{  }|@{  }l@{  }|l|l@{  }|l|l@{  }|}
        \hline
        \multirow{2}{*}{Video}  & \multirow{2}{*}{Resolution}   & \multirow{2}{*}{NI} & \multicolumn{2}{c|}{ Time/s } & \multicolumn{2}{c|}{Fm value}  \\
        \cline{4-7}
                                &                               &  & BRM &  IBRM &  BRM &  IBRM \\
        \hline
        \multirow{3}{*}{highway} & \multirow{3}{*}{$320 \times 240$}  & 1    & \textbf{1.2533}   & 1.4429  & 0.9554 & \textbf{0.9612}  \\
               &                  & 20                                      & 21.1842  & \textbf{2.2524}  & 0.9826 & \textbf{0.9828}  \\
            &                     & 50                                      & 52.7545  & \textbf{3.5712}  & \textbf{0.9911} & 0.9905  \\
           \hline
         \multirow{3}{*}{canoe} & \multirow{3}{*}{$320 \times 240$}    & 1  & \textbf{1.1496}   &  1.4215  &   \textbf{0.9552} & 0.9528 \\
               &                  & 20                                      & 19.9488  &  \textbf{2.2968}  &   \textbf{0.9535} & 0.9482 \\
            &                     & 50                                      & 50.4685  &  \textbf{3.5780}  &   0.9534 & \textbf{0.9453} \\
           \hline
         \multirow{3}{*}{wetSnow} &    \multirow{3}{*}{$720 \times 540$}     & 1    & 5.2601   & \textbf{1.6945}  & 0.7252   &  \textbf{0.7279}  \\
               &                  & 20                                              & 93.1602  & \textbf{5.8695}  & \textbf{0.7731}   &  0.7646  \\
            &                     & 50                                              & 234.7885 & \textbf{12.5385} & \textbf{0.7750}   &  0.7732  \\
            \hline
    \end{tabular}

    NI: Number of iterations \ BRM: Bayesian refinement model \ I\_BRM:Improved Bayesian refinement model.
    \vspace{-15pt}
\end{table}

\begin{table*}[!t]				
    \caption{Details of the network architecture of arithmetic distribution neural networks (ADNNs) and convolutional neural networks (CNNs), used for comparison to demonstrate the superiority of the proposed approach.}
\vspace{-5pt}
\label{network_architecture_compare}
\centering
    \begin{tabular}{|@{ }l@{ }|@{ }l@{ }|@{ }l@{ }|@{ }l@{ }|@{ }l@{ }|@{ }l@{ }|@{ }l@{ }|@{ }l@{ }|@{ }l@{ }|@{ }l@{ }|@{ }l@{ }|@{ }l|l@{ }|}
\hline
        \multicolumn{3}{|c@{ }|@{ }}{CNN$_1$ } &   \multicolumn{3}{c|@{ }}{CNN$_2$}  &   \multicolumn{3}{c|@{ }}{CNN$_3$}  &   \multicolumn{3}{@{ }c@{ }|}{CNN$_4$} & \multirow{2}{*}{Data Size} \\
        \cline{1-12}
        Type & Filters & Size &   Type & Filters & Size   &   Type & Filters & Size  &   Type & Filters & Size & \\
        \hline
        Input   &  &                &          &        &                   &    &  &   &    &  &                                                                       & B$\times 3 \times 201 \times 1$  \\
                &  &                &          &        &                   &    &  &                                                                 &   Conv  & 8   & $3 \times 1 \times 1$                                                                    &     \\
                &  &                &          &        &                                   &   Conv & 8    & $3 \times 1 \times 1$                   &   Conv  & 16  & $8 \times 1 \times 1$                                      & \\
                &  &                &          &        &                                   &   Conv & 16   & $8 \times 1 \times 1$                   &   Conv  & 32  & $16 \times 201 \times 1$                                       & \\
                &  &                & Conv     & 8      & $3 \times 1 \times 1$             &   Conv & 32   & $16 \times 201 \times 1$                &   Gelu  &     &                                      & \\
                &  &                & Conv     & 16     & $8 \times 201 \times 1$           &   Conv & 128  & $32 \times 1 \times 1$                  &   Conv  & 128 & $32 \times 1 \times 1$                            & \\
\hline
\end{tabular}
ProDis: product distribution layer \ SumDis: sum distribution layer \ Conv: Convolution \ Gelu: Gaussian Error Linear Unit \ Relu: Rectified Linear Unit \ NTP: Number of total parameters \ B: batch size
\vspace{-10pt}
\end{table*}
\begin{table*}[!t]				
    \caption{Quantitative comparison between CNNs and ADNN using Re, Pr, and Fm metrics based on real data.}
    \vspace{-5pt}
\label{tab_provscnn}
\centering
    \begin{tabular}{|l@{ \  \ }c@{ \  }c@{  \ \ }cc@{  \  }c@{ \ \  }c@{  \  }c@{ \ \  }c@{ \   }c@{ \   }c@{ \ \  }c@{ \  }c@{ \ }c@{\  }c@{ \  }c@{\ \ }|}
\hline
        \multirow{2}{*}{Videos} & \multicolumn{3}{c}{CNN$_1$} &  & \multicolumn{3}{c}{CNN$_2$} & & \multicolumn{3}{c}{CNN$_3$}   & & \multicolumn{3}{c|}{CNN$_4$}   \\
        \cline{2-4} \cline{6-8} \cline{10-12} \cline{14-16}
       & Re &Pr & Fm &  &  Re & Pr & Fm  &  &  Re & Pr & Fm    &  &  Re & Pr & Fm    \\
\hline
   highway         &   (0.9086 ,\   &  0.8133 ,\  &   0.8583 )   &  &    (0.8732  &  ,\ 0.7599  &  ,\ 0.8126 )   &  &    (0.9532  &  ,\  0.8017  &  ,\  0.8709 )         &  & (0.9085 &  ,\ 0.8805 &  ,\ 0.8943 ) \\  
   pedestrians     &   (0.9370 ,\   &  0.9404 ,\  &   0.9387 )   &  &    ({0.9536}  &  ,\ 0.9062  &  ,\ 0.9293 )   &  &    (0.9474  &  ,\  0.8942  &  ,\  0.9201 )       &  & (0.8963 &  ,\ 0.8518 &  ,\ 0.8735 ) \\  
   fountain01      &   (0.1404 ,\   &  0.1988 ,\  &   0.1646 )   &  &    ({0.6434}  &  ,\ 0.4435  &  ,\ 0.5250 )   &  &    (0.5944  &  ,\  0.5001  &  ,\  0.5432 )       &  & (0.6538 &  ,\ 0.4665 &  ,\ 0.5445 ) \\  
   canoe           &   (0.8134 ,\   &  0.8398 ,\  &   0.8264 )   &  &    ({0.9464}  &  ,\ 0.9532  &  ,\ 0.9498 )   &  &    (0.9439  &  ,\  {0.9663}i  &  ,\  0.9550 )    &  & (0.9422 &  ,\ 0.9560 &  ,\ 0.9490 ) \\  
   fountain02      &   (0.7289 ,\   &  0.8812 ,\  &   0.7978 )   &  &    ({0.7850}  &  ,\ 0.8584  &  ,\ 0.8201 )   &  &    (0.7670  &  ,\  0.9186  &  ,\  0.8360 )       &  & (0.7391 &  ,\ 0.9505 &  ,\ 0.8316 ) \\  
   peopleInShade   &   ({0.9893} ,\   &  0.2478 ,\  &   0.3963 )   &  &    (0.9054  &  ,\ 0.4649  &  ,\ 0.6144 )   &  &    (0.8941  &  ,\  {0.4758}  &  ,\  \textbf{0.6211} )   &  & (0.8921 &  ,\ 0.3713 &  ,\ 0.5243 ) \\
\hline

\end{tabular}
        \vspace{-10pt}
\end{table*}

\begin{table*}[htbp]				
    \caption{Quantitative comparison between CNNs and ADNN using Re, Pr, and Fm metrics based on seen and unseen videos.}
        \vspace{-5pt}
\label{tab_general_unseen}
\centering
    \begin{tabular}{|@{ }l@{ }|@{}l@{}l@{}|@{}l@{ }l@{}|@{ }c@{ }c@{ }c@{ }|c@{ }c@{ }c@{ }|c@{ }c@{ }c@{ }|c@{ }c@{ }c@{ }|}
\hline
                 \multicolumn{3}{|c}{Training Sets} &  \multicolumn{2}{@{}c}{Testing Sets}   &   \multicolumn{3}{c@{}}{CNN$_8$}  & \multicolumn{3}{c@{}}{CNN$_9$} & \multicolumn{3}{c@{}}{CNN$_{10}$} & \multicolumn{3}{c|@{}}{CNN$_{11}$} \\
         \cline{1-3} \cline{4-5} \cline{6-17}
        &        \ Category & Video  & \ Category & Video  & Re & Pr & Fm & Re & Pr & Fm & Re & Pr & Fm  & Re & Pr & Fm  \\
\hline
        \parbox[t]{2mm}{\multirow{8}{*}{\rotatebox[origin=c]{90}{ \ }}} &
            \multirow{4}{*}{ \tabincell{l}{Baseline \\
                                           Baseline}}  
        &   \multirow{4}{*}{\tabincell{l}{highway \\ 
                                                pedestrians}}
                                                 & \ Baseline & highway                 & (0.9343, & 0.7418, & 0.8270)  & (0.9321, & 0.9275, & 0.9298)  & (0.9341, & 0.9134, & 0.9237) & (0.9341, & 	0.9056, & 	0.9196)\\
        &                    &                   & \ Baseline & pedestrians             & (0.9574, & 0.8828, & 0.9186)  & (0.9280, & 0.8972, & 0.9123)  & (0.9362, & 0.9070, & 0.9213) & (0.9486, & 	0.8929, & 	0.9199)\\
        &                    &                   & \ Baseline & office                  & (0.9314, & 0.7209, & 0.8128)  & (0.8751, & 0.7307, & 0.7964)  & (0.8848, & 0.7318, & 0.8011) & (0.8784, & 	0.7377, & 	0.8019)\\
        &                    &                   & \ Baseline & PETS2006                & (0.9076, & 0.7932, & 0.8465)  & (0.8872, & 0.8183, & 0.8514)  & (0.8885, & 0.7979, & 0.8408) & (0.8946, & 	0.8051, &   0.8475)\\
\hline
\end{tabular}
        \vspace{-10pt}
\end{table*}

\begin{table*}[!t]				
    \caption{Details of the network architecture for arithmetic distribution neural networks (ADNNs) used in ablation study.}
\vspace{-5pt}
\label{network_architecture_adnn}
\centering
    \begin{tabular}{|@{ }l@{ }|@{ }l@{ }|@{ }l@{ }|@{ }l@{ }|@{ }l@{ }|@{ }l@{ }|@{ }l@{ }|@{ }l@{ }|@{ }l@{ }|@{ }l@{ }|@{ }l@{ }|@{ }l|l@{ }|}
\hline
        \multicolumn{3}{|c@{ }|@{ }}{ADNN$_2$ } &   \multicolumn{3}{c|@{ }}{ADNN$_3$}  &   \multicolumn{3}{c|@{ }}{ADNN$_4$}  &   \multicolumn{3}{@{ }c@{ }|}{ADNN$_5$} & \multirow{2}{*}{Data Size} \\
        \cline{1-12}
        Type & Filters & Size &   Type & Filters & Size   &   Type & Filters & Size  &   Type & Filters & Size & \\
        \hline
        Input   &  &                            &          &        &                                   &        &      &                                         &    &  &                         & B$\times 3 \times 201 \times 1$  \\
        ProDis  &2   & $3 \times 201 \times 1$    & ProDis  &2   & $3 \times 11 \times 1$      &  ProDis  & 1   & $3 \times 21 \times 1$                & ProDis   & 2   & $3 \times 21 \times 1$   & \\
        SumDis  &2   & $3 \times 201 \times 1$    & SumDis  &2   & $3 \times 11 \times 1$      &  SumDis  & 1   & $3 \times 21 \times 1$                & SumDis   & 2   & $3 \times 21 \times 1$   & \\
        Conv    &1   & $3 \times 1 \times 2$      & Conv    &1   & $3 \times 1 \times 2$       &  Conv    & 1   & $3 \times 1 \times 1$                 & Conv     & 1   & $3 \times 1 \times 2$    & \\
        Conv    &512 & $1 \times 201 \times 1$    & Conv    &512 & $1 \times 11 \times 1$      &  Conv    & 512 & $1 \times 21 \times 1$                & Conv     & 512 & $1 \times 21 \times 1$   & B$\times 512 \times 1 \times 1$ \\
        Relu    &    &                            & Relu    &    &                             &  Relu    &     &                                       & Relu     &     &                          & \\
\hline
    \end{tabular}

ProDis: product distribution layer \ SumDis: sum distribution layer \ Conv: Convolution \ Gelu: Gaussian Error Linear Unit \ Relu: Rectified Linear Unit \ NTP: Number of total parameters \ B: batch size
    \vspace{-5pt}
\end{table*}

\begin{table*}[htbp]				
    \caption{Quantitative comparisons between ADNN$_{2-9}$ for ablation study using Re, Pr, and Fm metrics based on seen and unseen videos.}
        \vspace{-5pt}
\label{tab_general_unseen_adnn}
\centering
    \begin{tabular}{|@{ }l@{ }|@{}l@{}l@{}|@{}l@{ }l@{}|@{ }c@{ }c@{ }c@{ }|c@{ }c@{ }c@{ }|c@{ }c@{ }c@{ }|c@{ }c@{ }c@{ }|}
\hline
                 \multicolumn{3}{|c}{Training Sets} &  \multicolumn{2}{@{}c}{Testing Sets}   &   \multicolumn{3}{c@{}}{ADNN$_2$}  & \multicolumn{3}{c@{}}{ADNN$_3$} & \multicolumn{3}{c@{}}{ADNN$_4$} & \multicolumn{3}{c|@{}}{ADNN$_5$} \\
         \cline{1-3} \cline{4-5} \cline{6-17}
        &        \ Category & Video  & \ Category & Video  & Re & Pr & Fm & Re & Pr & Fm & Re & Pr & Fm  & Re & Pr & Fm  \\
\hline
        \parbox[t]{2mm}{\multirow{8}{*}{\rotatebox[origin=c]{90}{ \ }}} &
            \multirow{4}{*}{ \tabincell{l}{Baseline \\
                                           Baseline}}  
        &   \multirow{4}{*}{\tabincell{l}{highway \\ 
                                                pedestrians}}
                                                 & \ Baseline & highway                 & (0.9303, & 0.9378, & 0.9340)  & (0.9296, &  0.8883, &  0.9085)  & (0.9307, &    0.8764, &  0.9027) & (0.9284, &  0.8920, &  0.9098)\\
        &                    &                   & \ Baseline & pedestrians             & (0.9710, & 0.9214, & 0.9455)  & (0.9491, &  0.9172, &  0.9329)  & (0.9441, &    0.9096, &  0.9265) & (0.9407, &  0.9344, &  0.9375)\\
        &                    &                   & \ Baseline & office                  & (0.9250, & 0.8720, & 0.8977)  & (0.8680, &  0.7487, &  0.8040)  & (0.8678, &    0.7769, &  0.8199) & (0.8696, &  0.7859, &  0.8256)\\
        &                    &                   & \ Baseline & PETS2006                & (0.9081, & 0.8137, & 0.8583)  & (0.8945, &  0.7993, &  0.8442)  & (0.8776, &    0.8158, &  0.8455) & (0.8701, &  0.8378, &  0.8537)\\
\hline
\end{tabular}
        \vspace{-10pt}
\end{table*}

\begin{table*}[!t]				
    \caption{Quantitative evaluation of the proposed approach for unseen videos on the CDnet2014 \cite{2014_CDN_6910011} dataset, using the Fm metric.}
\vspace{-5pt}
\label{tab_unseen_CDnet}					
\centering
    \begin{tabular}{|l@{ }|l@{ }|c@{  }c@{  }c@{  }c@{  }c@{  }c@{  }c@{  }c@{  }c@{  }c@{    }c@{   }|c@{  }|}
\hline
        \multicolumn{2}{|c|}{Approach}  &  \ Baseline\   &  \ Dyn. Bg.\  &   \  Cam. Jitt.\   &     \ Int. Mot.\  &  \ \ Shadow\ \   &  \ Ther.\   &  \ Bad Wea.\   &  \ Low Fr.\  & \ Nig. Vid.\   &    \ \ PTZ\ \     &      \ \ Turbul.\   &     \ Overall\    \\
\hline
  \parbox[t]{2mm}{\multirow{16}{*}{\rotatebox[origin=c]{90}{   \ \ \   }}}     &      IUTIS-5\cite{2017_ICIAP_combing}    & 0.9567   & 0.8902   & 0.8332   & 0.7296   & 0.9084   & 0.8303   & 0.8248   & {0.7743}   & 0.5290   & 0.4282   & 0.7836   & 0.7717   \\
        &           PAWCS\cite{2016_TIP_7539354}   & 0.9397   & 0.8938             & 0.8137   & 0.7764   & 0.8913   & 0.8324   & 0.8152   & 0.6588   & 0.4152   & 0.4615   & 0.6450   & 0.7403   \\
        &             BSUV\cite{2020_WACV_Tezcan}  & \textbf{0.9693} & 0.7967 & 0.7743 & 0.7499 & 0.9233 & 0.8581 & 0.8713 & 0.6797 & \textbf{0.6987} & 0.6282 & 0.7051   & 0.7868   \\
        &             BSUV2.0\cite{2021_ACCEE_9395443}  & 0.9620 & 0.9057 & \textbf{0.9004} & \textbf{0.8263} & \textbf{0.9562} & \textbf{0.8932} & \textbf{0.8844} & \textbf{0.7902} & 0.5857 & \textbf{0.7037} & \textbf{0.8174} & \textbf{0.8387} \\
        &             FgSN\cite{2018_PRL_LIM2018256}  & 0.6926 & 0.3634 & 0.4266 & 0.2002 & 0.5295 & 0.6038 & 0.3277 & 0.2482 & 0.2800 & 0.3503 & 0.0643  & 0.3715    \\
        \hline
        &    ADNN-IB$_{\text{U4fs}}$   &  0.9522   &   \textbf{0.9166}        &   0.8245   &  0.6978   & 0.9054   & 0.8058    & 0.8245    & 0.7028   & 0.4872   &  0.2931    & 0.7620    & 0.7451   \\
\hline
\end{tabular}
        \vspace{-15pt}
\end{table*}

\begin{table}[!t]				
    \caption{Quantitative evaluation of the proposed approach for unseen videos on the LASIESTA \cite{CUEVAS2016103} dataset, using the Fm metric.}
\vspace{-5pt}
\label{tab_unseen_LAS}					
\centering
    \begin{tabular}{|@{ }l@{}|@{}c@{}|@{}c@{ }c@{ }c@{ }c@{}|}
\hline
        Videos &  ADNN-IB$_{\text{C2fs}}$ & BSUV2.0\cite{2021_ACCEE_9395443} & 3DCD\cite{2020_TIP_9263106}  & FgSN\cite{2018_PRL_LIM2018256} & CueV2\cite{2018_PR_BERJON}  \\
\hline
        I\_SI\_02    & \textbf{0.91}  & 0.89  & 0.86   & 0.53   &   0.84   \\
        I\_CA\_02    & \textbf{0.79}  & 0.60  & 0.49   & 0.58   &   0.78   \\
        I\_OC\_02    & \textbf{0.95}  & \textbf{0.95}  & 0.93   & 0.25   &   0.86   \\
        I\_IL\_02    & 0.53  & \textbf{0.89}  & 0.85   & 0.41   &   0.65   \\
        I\_MB\_02    & 0.88  & 0.76  & 0.79   & 0.63   &   \textbf{0.92}   \\
        I\_BS\_02    & \textbf{0.92}  & 0.69  & 0.87   & 0.25   &   0.62   \\
        O\_CL\_02    & \textbf{0.97}  & 0.89  & 0.87   & 0.54   &   0.90   \\
        O\_RA\_02    & \textbf{0.97}  & 0.93  & 0.87   & 0.54   &   0.89   \\
        O\_SN\_02    & \textbf{0.80}  & 0.70  & 0.49   & 0.05   &   0.63   \\
        O\_SU\_02    & 0.87  & \textbf{0.91}  & 0.83   & 0.29   &   0.77   \\
\hline                               
        Average      & \textbf{0.86}  & 0.82  & 0.79   & 0.41   &   0.79   \\
\hline
\end{tabular}
        \vspace{-15pt}
\end{table}

\section{Experiments}
\subsection{Comparisons with Convolutional Neural Networks}
\label{sec_comp_CNN}
In this section, we demonstrate the superiority of the proposed arithmetic distribution neural network (ADNN) compared to the convolutional neural network (CNN).
Arithmetic distribution layers are proposed to serve as a better substitute for the convolutional layer. Thus, the proposed ADNN is better than convolutional neural networks in distribution classification.
In order to demonstrate this, we devise 2 arithmetic distribution neural networks (ADNN$_{1-2}$) to compare with 14 traditional convolutional neural networks (CNN$_{1-14}$).
Details on the architecture of ADNN$_{1-2}$ and CNN$_{1-14}$ are shown in \reftab{network_architecture_compare}.
CNN$_{1-3}$ are devised by replacing the arithmetic distribution layers in ADNN$_1$ by several convolutional layers.
CNN$_{4-7}$ are devised by inserting extra nonlinear activation functions, such as Rectified Linear Unit (Relu) or Gaussian Error Linear Unit (Gelu), into CNN$_3$ to handle the nonlinear data.
Similarly, CNN$_{8-10}$ are devised by replacing the arithmetic distribution layers in ADNN$_2$ with fully connected layers,
and CNN$_{11-14}$ are devised by inserting extra Relu and Gelu layers into CNN$_{10}$.
During the comparisons between ADNNs and CNNs,
one ground-truth frame from training videos is extracted for training, and all training and testing settings of CNNs and ADNNs are the same, including the learning rate, training algorithms, maximum number of training epochs, random number seed as well as the input and output of networks.
For evaluation, the Re (Recall), Pr (Precision), and Fm (F-measure) metrics are used.
\begin{table*}[!t]				
    \caption{Quantitative evaluation of the proposed approach for seen and unseen videos on the LASIESTA \cite{CUEVAS2016103} dataset, using the Fm metric.}
    \vspace{-5pt}
\label{tab_fg_LAS}					
\centering
    \begin{tabular}{|@{   }l@{   }|@{   }c@{   }|@{   }c@{   }c@{   }c@{   }c@{   }|c@{}c@{ }|c@{ }c@{ }c@{  }|}
\hline
        Videos & ADNN-IB$_{\text{L3fs}}$ & D-DPDL\cite{2019_TCSVT_8892609} &     CueV2\cite{2018_PR_BERJON} & Hai\cite{2014_TPAMI_6678500}  & Cue\cite{CUEVAS2013616} &    ADNN-IB$_{\text{C2fs}}$ &  ADNN-IB$_{\text{c8fm}}$    & BSUV2.0\cite{2021_ACCEE_9395443}  & 3DCD-55\cite{2020_TIP_9263106} & MSFS-55\cite{lim2020learning} \\
\hline
        I\_SI\_01    & \textbf{0.9764}  & 0.9596     & 0.9208   & 0.9622   &  0.8143  &  \multirow{2}{*}{0.9293} &   \multirow{2}{*}{0.9335} &  \multirow{2}{*}{0.9200} &   \multirow{2}{*}{0.8700}  &   \multirow{2}{*}{0.3900} 	\\
        I\_SI\_02    & \textbf{0.9309}  & 0.8687     & 0.8403   & 0.8130   &  0.7576  &                          &                           &                          &                            &                              \\
        I\_CA\_01    & \textbf{0.9807}  & 0.9309     & 0.9062   & 0.9220   &  0.8424  &  \multirow{2}{*}{0.8225} &   \multirow{2}{*}{0.7316} &  \multirow{2}{*}{0.6800} &   \multirow{2}{*}{0.8200}  &   \multirow{2}{*}{0.4000} 	\\
        I\_CA\_02    & \textbf{0.9201}  & 0.8850     & 0.7826   & 0.8656   &  0.6296  &                          &                           &                          &                            &                              \\
        I\_OC\_01    & \textbf{0.9783}  & 0.9710     & 0.7013   & 0.8920   &  0.8274  &  \multirow{2}{*}{0.9600} &   \multirow{2}{*}{0.9481} &  \multirow{2}{*}{0.9600} &   \multirow{2}{*}{0.9100}  &   \multirow{2}{*}{0.3700} 	\\
        I\_OC\_02    & \textbf{0.9735}  & 0.9677     & 0.8600   & 0.9526   &  0.8781  &                          &                           &                          &                            &                              \\
        I\_IL\_01    & 0.7702  & 0.7161     & 0.6452   & \textbf{0.8861}   &  0.7966  &  \multirow{2}{*}{0.5216} &   \multirow{2}{*}{0.4869} &  \multirow{2}{*}{0.8800} &   \multirow{2}{*}{0.9200}  &   \multirow{2}{*}{0.3500} 	\\
        I\_IL\_02    & 0.7620  & \textbf{0.8972}     & 0.6523   & 0.8122   &  0.7864  &                          &                           &                          &                            &                              \\
        I\_MB\_01    & \textbf{0.9874}  & 0.9699     & 0.9543   & 0.9816   &  0.7779  &  \multirow{2}{*}{0.9272} &   \multirow{2}{*}{0.9262} &  \multirow{2}{*}{0.8100} &   \multirow{2}{*}{0.8900}  &   \multirow{2}{*}{0.6400} 	\\
        I\_MB\_02    & \textbf{0.9731}  & 0.9195     & 0.9204   & 0.7064   &  0.6797  &                          &                           &                          &                            &                              \\
        I\_BS\_01    & \textbf{0.9787}  & 0.8371     & 0.7132   & 0.6285   &  0.5065  &  \multirow{2}{*}{0.9216} &   \multirow{2}{*}{0.8884} &  \multirow{2}{*}{0.7700} &   \multirow{2}{*}{0.7200}  &   \multirow{2}{*}{0.3600} 	\\
        I\_BS\_02    & \textbf{0.9626}  & 0.6178     & 0.6156   & 0.7333   &  0.6607  &                          &                           &                          &                            &                              \\
        O\_CL\_01    & \textbf{0.9840}  & 0.9792     & 0.9508   & 0.6946   &  0.9280  &  \multirow{2}{*}{0.9594} &   \multirow{2}{*}{0.9534} &  \multirow{2}{*}{0.9300} &   \multirow{2}{*}{0.8700}  &   \multirow{2}{*}{0.4100} 	\\
        O\_CL\_02    & 0.9788  & \textbf{0.9800}     & 0.9045   & 0.9588   &  0.8995  &                          &                           &                          &                            &                              \\
        O\_RA\_01    & \textbf{0.9896}  & 0.9072     & 0.8453   & 0.8225   &  0.7462  &  \multirow{2}{*}{0.8668} &   \multirow{2}{*}{0.8744} &  \multirow{2}{*}{0.9400} &   \multirow{2}{*}{0.9000}  &   \multirow{2}{*}{0.3500} 	\\
        O\_RA\_02    & \textbf{0.9839}  & 0.9803     & 0.8886   & 0.9590   &  0.8699  &                          &                           &                          &                            &                              \\
        O\_SN\_01    & \textbf{0.9733}  & 0.9690     & 0.9317   & 0.3054   &  0.8214  &  \multirow{2}{*}{0.8300} &   \multirow{2}{*}{0.8327} &  \multirow{2}{*}{0.8400} &   \multirow{2}{*}{0.6900}  &   \multirow{2}{*}{0.3100} 	\\
        O\_SN\_02    & \textbf{0.9562}  & 0.9341     & 0.6256   & 0.0426   &  0.0895  &                          &                           &                          &                            &                              \\
        O\_SU\_01    & \textbf{0.9186}  & 0.9065     & 0.6774   & 0.8115   &  0.6527  &  \multirow{2}{*}{0.8715} &   \multirow{2}{*}{0.8829} &  \multirow{2}{*}{0.7900} &   \multirow{2}{*}{0.8500}  &   \multirow{2}{*}{0.3700} 	\\
        O\_SU\_02    & \textbf{0.9413}  & 0.9388     & 0.7669   & 0.9021   &  0.8074  &                          &                           &                          &                            &                              \\
\hline                                                                                              
        Average      & \textbf{0.9460}  & 0.9068     & 0.8051   & 0.7826   &  0.7386  &  0.8610 & 0.8459 & 0.8500 &  0.8400      & 0.4000           \\
\hline
\end{tabular}
        \vspace{-10pt}
\end{table*}

From the quantitative comparisons shown in \reftab{tab_provscnn}, we can conclude that the results of CNN$_3$ are better than CNN$_2$ which are better than CNN$_1$, since when the training data is fixed, a network with more parameters is supposed to have better learning ability.
In addition, the results of CNN$_3$ are better than the ones of CNN$_{4-7}$ which are devised by adding Relu and Gelu layers into CNN$_3$. This is because Relu or Gelu drops several entires of input vectors which may result in losing useful information.
In contrast, the proposed ADNN$_1$ achieves better results on the average Fm value compared to CNN$_{3-7}$, which uses around 100 times more parameters than the proposed ADNN.
This clearly demonstrates that the block consisting of arithmetic distribution layers has better distribution learning ability than the one consisting of traditional convolutional layers attached or not attached with Relu or Gelu.
Moreover, in order to further compare the proposed arithmetic distribution layers with fully connected layers, which has better ability than convolutional layers to learn global information, the comparisons between ADNN$_2$ and CNN$_{8-14}$ are presented in \reftab{tab_general_unseen}.
In particular, since the learning ability of CNN and ADNN are improved by increasing of the number of parameters,
both the seen and unseen videos are used for evaluation to obtain the quantitative results under challenging conditions.
As shown in \reftab{tab_general_unseen}, CNN$_6$ is the largest network with 2.9 million parameters and it is constructed purely by fully connected layers, and CNN$_{11-14}$ are devised by inserting Relu and Gelu in these fully connected layers.
However, the proposed ADNN$_2$ still achieves better results than CNN$_{8-14}$ and the number of parameters in ADNN$_2$ is only around 0.1 million.
Thus, it is fair to claim that the proposed arithmetic distribution layers are better than convolutional layers in distribution classification tasks.

\subsection{Ablation Study}
\label{sec_abla}
Although the comparisons between CNNs and the proposed ADNNs have been proposed in \refsec{sec_comp_CNN},
in which the benefit of using the proposed arithmetic distributions layers is demonstrated,
it is still not clear how the kernel size, the filter number as well as the depth of arithmetic distribution layers contribute to the results.
Thus, in this section, we devise another 7 ADNNs (ADNN$_{3-9}$), which are revised from ADNN$_2$, to compare with ADNN$_2$ for ablation study.
The details of network architectures of ADNN$_{2-9}$ are shown in \reftab{network_architecture_adnn}, and their quantitative comparisons are shown in \reftab{tab_general_unseen_adnn}.
During the evaluations of ADNN$_{3-9}$, the training and testing setting are the same as those of ADNN$_2$.

Compared to ADNN$_2$, ADNN$_{6-9}$ have more parameters,
but their overall Fm values are close to those of ADNN$_2$.
This demonstrates that ADNN$_2$ has enough learning ability to handle the histograms of temporal pixels for background subtraction.
In particular, ADNN$_6$ and ADNN$_7$ are devised by increasing the filter number and the width of arithmetic distribution layers, respectively.
ADNN$_8$ is devised by increasing the filter number as well as the width of the arithmetic distribution layer.
ADNN$_9$ is devised by increasing the depth of arithmetic distribution layers.
Note that the overall Fm value of ADNN$_9$ is a little lower than ADNN$_2$, this decline may be from the accumulated error or the over-fitting problem, since ADNN$_9$ has a deeper architecture.
In contrast, the number of parameters in ADNN$_{3-5}$ is less than the one in ADNN$_2$, and their Fm values are thus lower.
In particular, ADNN$_5$ is devised by reducing the width of arithmetic distribution layers from 201 to 21, and the overall Fm value of ADNN$_5$ is 0.8492. This demonstrates the importance of the width of the proposed arithmetic distribution layers.
Furthermore, the number of filters is reduced to 1 in ADNN$_4$,
but the overall Fm value of ADNN$_4$ is close to ADNN$_5$.
Thus, compared to the number of filters, the width of the proposed arithmetic distribution layers is more important.
In order to further demonstrate this point,
ADNN$_3$ is devised to compare with ADNN$_4$ and ADNN$_5$.
ADNN$_3$ has the same number of filters as ADNN$_5$, but the width of arithmetic distribution layers is smaller than those of ADNN$_4$ and ADNN$_5$.
As the results show in \reftab{tab_general_unseen_adnn},
the overall Fm value of ADNN$_3$ has an obvious decline compared to ADNN$_4$, which has a value similar to ADNN$_5$.
This again demonstrates the importance of the width of the proposed arithmetic distribution layers.
Furthermore, as the forward pass and backpropagation formulae shown in \refequ{eq_product_distribution} and \refequ{eq_prodis_back}, respectively,
the domain of $i$, which represents the width of the layers, is theoretically $[-\infty, \infty]$.
However, for real applications we have to use a finite domain to replace the infinite domain.
Thus, the accuracy of the arithmetic distribution operations is actually limited by the width of the proposed arithmetic distribution layers. This is the main reason why the layer width is important.

\subsection{Comparisons with state-of-the-art methods}
\label{sec_evaluation}
The proposed approach has three good properties for use in real applications.
1) \emph{Generality}: the proposed approach is effective for unseen videos;
2) \emph{Efficiency}: only limited ground-truth frames are required to generate promising results;
3) \emph{Universality}: one network can be trained for all videos.
In this section, these three properties are demonstrated through comparisons with state-of-the-art methods including unsupervised methods, deep learning methods for seen videos and unseen videos on CDnet2014 \cite{2014_CDN_6910011} LASIESTA \cite{CUEVAS2016103} and SBMI2015 \cite{10.1007/978-3-319-23222-5_57} datasets.

The deep learning networks compared include DeepBS\cite{2018_PR_BABAEE2018635}, GuidedBS\cite{2018_ICME_8486556}, CNN-SFC\cite{2019_JEI_bgconv}, D-DPDL \cite{2019_TCSVT_8892609}, 3DCD\cite{2020_TIP_9263106}, FgSN\cite{2018_PRL_LIM2018256}, MSFS\cite{lim2020learning}, DVTN \cite{2020_TCSVT_9281081} and BSUV\cite{2020_WACV_Tezcan,2021_ACCEE_9395443}.
In particular 3DCD\cite{2020_TIP_9263106} and BSUV\cite{2020_WACV_Tezcan, 2021_ACCEE_9395443} are effective for unseen videos.
After the rise of deep learning networks in the background subtraction field, the fairness of comparisons between deep learning methods has been a concern.
It is commonly accepted that the quantity of training data and the number of parameters in a network have significant and direct contributions to the performance of various methods \cite{krizhevsky2012imagenet}. However, the assumptions on the training data, numbers of parameters in the network, and the utilization of pre-trained networks in these methods are completely different.
In order to propose fair comparisons, the proposed ADNN is trained and tested under 6 conditions to propose different evaluation results which are named ADNN-IB$_{\text{U4fs}}$, ADNN-IB$_{\text{C2fs}}$, ADNN-IB$_{\text{L3fs}}$, ADNN-IB$_{\text{S2fs}}$, ADNN-IB$_{\text{c8fm}}$ and ADNN-IB$_{\text{C20fs}}$ for comparisons with various state-of-the-art methods.

ADNN-IB$_{\text{U4fs}}$ is trained following the partition of training and testing videos proposed in BSUV \cite{2020_WACV_Tezcan}.
During training, 4 ground-truth frames from every training video and 4 binary masks of testing videos provided by IUTIS-5 \cite{2017_ICIAP_combing} are mixed and used for training.
Note that BSUV also manually extracted hundreds of frames from testing videos to generate reference images which can be used as background instances to train the proposed approach.
ADNN-IB$_{\text{C2fs}}$ is trained by a mixture of 2 ground-truth frames from every video of the CDnet2014 \cite{2014_CDN_6910011} dataset.
ADNN-IB$_{\text{c8fm}}$ is trained by a mixture of 8 down-sampling ground-truth frames from every video of the CDnet2014 \cite{2014_CDN_6910011} dataset.
During the down-sampling procedure, only 25\% of labels from the ground-truth frames are kept for training. 
Thus, the total number of training instances of ADNN-IB$_{\text{c8fm}}$ and ADNN-IB$_{\text{C2fs}}$ are actually the same, but the training set of ADNN-IB$_{\text{c8fm}}$ includes more temporal information.
Both ADNN-IB$_{\text{C2fs}}$ and ADNN-IB$_{\text{c8fm}}$ are tested on CDnet2014 \cite{2014_CDN_6910011}, LASIESTA \cite{CUEVAS2016103} and SBMI2015 \cite{10.1007/978-3-319-23222-5_57} datasets.
During the evaluations of ADNN-IB$_{\text{c8fm}}$ and ADNN-IB$_{\text{C2fs}}$, only one network is trained and the parameters of the network are thus fixed for all testing videos.

ADNN-IB$_{\text{L3fs}}$, ADNN-IB$_{\text{S2fs}}$ and ADNN-IB$_{\text{C20fs}}$ are evaluated on LASIESTA \cite{CUEVAS2016103}, SBMI2015 \cite{10.1007/978-3-319-23222-5_57} and CDnet2014 \cite{2014_CDN_6910011} datasets respectively.
During the evaluations, 3, 2, and 20 ground-truth frames from videos of corresponding datasets are used for training, and the remaining frames of the same videos are used for testing. This means that various networks are trained for different videos to generate the results of ADNN-IB$_{\text{L3fs}}$, ADNN-IB$_{\text{S2fs}}$ and ADNN-IB$_{\text{C20fs}}$.
The training frames only take less than 1\% of ground-truth frames of the corresponding datasets and all ground-truth frames used for training are randomly selected.

\subsubsection{Generality}
The distributions of temporal pixels are relatively independent of the scene information, demonstrating that the proposed approach is generalizable. 
In order to demonstrate this, ADNN-IB$_{\text{U4fs}}$, ADNN-IB$_{\text{C2fs}}$ and ADNN-IB$_{\text{c8fm}}$ are trained and evaluated.
The quantitative results of ADNN-IB$_{\text{U4fs}}$ are shown in \reftab{tab_unseen_CDnet}.
ADNN-IB$_{\text{U4fs}}$ achieves the best results in the ``Dyn. Bg.'' category.
In this category, single background images are not enough to describe background information.
In contrast, distributions for temporal pixels have better description ability.
This is main reason why ADNN-IB$_{\text{U4fs}}$ has better results than  BSUV2.0 and IUTIS-5.
Unfortunately, the overall results of ADNN-IB$_{\text{U4fs}}$ can only be considered as promising during comparisons with BSUV and BSUV2.0.
However, such comparisons may not be fair to the proposed approach, since BSUV extracted 200 ground-truth frames from every training video, and BSUV2.0 also utilized synthetic ground-truth frames to train their network.
The synthetic ground-truth frames used in BSUV2.0 are very close to the ground-truth of testing videos, since they are generated by the fusion of manually selected frames from the testing video and moving objects segmented from other videos with the help of ground-truth frames, and videos from the same dataset are correlated to each other. However, when the training videos and testing videos are extracted from different datasets,
the contribution of the quantity of the training set is reduced due to the lower correlation between training videos and testing videos.
Thus, ADNN-IB$_{\text{C2fs}}$ and ADNN-IB$_{\text{c8fm}}$ are trained to compare with BSUV2.0 and 3DCD across datasets.
The quantitative results of ADNN-IB$_{\text{C2fs}}$ on several videos from LASIESTA  are shown in \reftab{tab_unseen_LAS}.
The LASIESTA dataset contains 20 videos from indoor and outdoor scenes.
In addition, it also includes videos with illumination changes (I\_IL) or camouflage (I\_CA), which are challenging scenes for background subtraction.
However, ADNN-IB$_{\text{C2fs}}$ achieves the best results compared to BSUV2.0 and 3DCD.
In particular, not only is the proposed ADNN-IB$_{\text{C2fs}}$, but also BSUV2.0 and 3DCD are trained on the CDnet2014  dataset and tested on the LASIESTA  dataset. This demonstrates the superiority of the generality of the proposed approach across different datasets. Such superiority is also demonstrated by the quantitative results shown in\reftab{tab_fg_LAS}.
The proposed ADNN-IB$_{\text{C2fs}}$ achieves better results than BSUV2.0 and 3DCD on the entire LASIESTA dataset. Thus, the ability of the proposed approach to generalize is demonstrated.
\begin{table*}[!t]				
    \caption{Quantitative evaluation of the proposed approach for seen and unseen videos on the SBMI2015 dataset, using the Fm metric.}
    \vspace{-5pt}
\label{tab_unseen_SBMI2015}					
\centering
    \begin{tabular}{|l@{ }|l@{  }c@{   }c@{   }c@{   }c@{   }c@{   }c@{   }c@{   }c@{   }c@{   }c@{    }c@{   }c@{   }|c@{   }|}

\hline
        Approach & Board  &   Cand.   &  CAVIAR1  &  CAVIAR2   &   CaVig. &  Foliage   &  HallA.  &  HighwayI   &  HighwayII  & HumanB.  &  IBMtest2     &   PeopleA.   &   Snellen  & Overall \\
        \hline
        Vibe\cite{Barnich2011_2011_TIP}                        & 0.7377 & 0.5020 & 0.8051 & 0.7347 & 0.3497 & 0.5539 & 0.6017 & 0.4150 & 0.5554 & 0.4268 & 0.7001 & 0.6111 & 0.3083 & 0.5617     \\
        RPCA\cite{kang2015robust}                        & 0.5304 & 0.4730 & 0.4204 & 0.1933 & 0.4720 & 0.4617 & 0.4525 & 0.5733 & 0.7335 & 0.5765 & 0.6714 & 0.3924 & 0.4345 & 0.4911     \\
        SuBSENSE \cite{2015_TIP_6975239}                   & 0.6588 & 0.6959 & 0.8783 & 0.8740 & 0.4080 & 0.1962 & 0.7559 & 0.5073 & 0.8779 & 0.8560 & 0.9281 & 0.4251 & 0.2467 & 0.6391     \\
        FgSN-M-55\cite{2018_PRL_LIM2018256}               & 0.8900 & 0.2100 & 0.7000 & 0.0500 & 0.5700 & \textbf{0.9100} & 0.7100 & 0.7500 & 0.3100 & 0.8300 & 0.8300 & \textbf{0.9000} & 0.5200 & 0.6300 \\
        MSFS-55\cite{lim2020learning}                     & 0.9100 & 0.2600 & 0.5700 & 0.0800 & 0.5700 & 0.8000 & 0.5200 & 0.8200 & 0.5800 & 0.6100 & 0.6000 & 0.8700 & 0.6800 & 0.6100 \\
        3DCD-55\cite{2020_TIP_9263106}                       & 0.8300 & 0.3500 & 0.7900 & 0.5600 & 0.4800 & 0.6900 & 0.5800 & 0.7300 & 0.7700 & 0.6500 & 0.7000 & 0.7800 & 0.7600 & 0.6700   \\
        \hline
         ADNN-IB$_{\text{C2fs}}$   & 0.4680 & 0.5981 & 0.7801 & 0.8646 & 0.6994 & 0.1420 & 0.8176 & 0.7376 & 0.9786 & 0.9329 & 0.8967 & 0.3893 & 0.0472 & 0.6425 \\
        ADNN-IB$_{\text{C8fm}}$   & 0.4527 & 0.5222 & 0.9169 & 0.8429 & 0.7259 & 0.0722 & 0.8200 & 0.7493 & 0.9827 & 0.9431 & 0.9348 & 0.3071 & 0.0445 & 0.6396  \\
        ADNN-IB$_{\text{S2fs}}$    & \textbf{0.9421} & \textbf{0.9242} & \textbf{0.9550} & \textbf{0.8865} & \textbf{0.9589} & 0.7528 & \textbf{0.9151} & \textbf{0.8689} & \textbf{0.9854} & \textbf{0.9525} & \textbf{0.9548} & 0.7108 & \textbf{0.7893} & \textbf{0.8920}  \\

\hline
\end{tabular}
        \vspace{-10pt}
\end{table*}
\begin{table*}[!t]				
    \caption{Quantitative evaluation of the proposed approach for seen videos on CDnet2014 \cite{2014_CDN_6910011} dataset, using Fm metric.}
    \vspace{-5pt}
\label{tab_fg}					
\centering
    \begin{tabular}{|l@{ }|l@{ }|c@{  }c@{  }c@{  }c@{  }c@{  }c@{  }c@{  }c@{  }c@{  }c@{    }c@{   }|c@{  }|}

\hline
        \multicolumn{2}{|c|}{Approach}  &  \ Baseline\   &  \ Dyn. Bg.\  &   \  Cam. Jitt.\   &     \ Int. Mot.\  &  \ \ Shadow\ \   &  \ Ther.\   &  \ Bad Wea.\   &  \ Low Fr.\  & \ Nig. Vid.\   &    \ \ PTZ\ \     &      \ \ Turbul.\   &     \ Overall\    \\
\hline
  \parbox[t]{2mm}{\multirow{16}{*}{\rotatebox[origin=c]{90}{   \ \ \   }}}     &      IUTIS-5\cite{2017_ICIAP_combing}    & 0.9567   & 0.8902   & 0.8332   & 0.7296   & 0.9084   & 0.8303   & 0.8248   & {0.7743}   & 0.5290   & 0.4282   & 0.7836   & 0.7717   \\
     &             MBS\cite{2017_TIP_7904604}   & 0.9287   & 0.7915             & 0.8367   & 0.7568   & 0.7968   & 0.8194   & 0.7980   & 0.6350   & 0.5158   & 0.5520   & 0.5858   & 0.7288   \\
     &           PAWCS\cite{2016_TIP_7539354}   & 0.9397   & 0.8938             & 0.8137   & 0.7764   & 0.8913   & 0.8324   & 0.8152   & 0.6588   & 0.4152   & 0.4615   & 0.6450   & 0.7403   \\
     &         ShareM\cite{2015_ICME_7177419}   & 0.9522   & 0.8222             & 0.8141   & 0.6727   & 0.8898   & 0.8319   & 0.8480   & 0.7286   & 0.5419   & 0.3860   & 0.7339   & 0.7474   \\
     &        SuBSENSE\cite{2015_TIP_6975239}   & 0.9503   & 0.8177             & 0.8152   & 0.6569   & 0.8986   & 0.8171   & 0.8619   & 0.6445   & 0.5599   & 0.3476   & 0.7792   & 0.7408   \\
     &        WeSamBE\cite{2017_TCSVT_7938679}  & 0.9413   & 0.7440             & 0.7976   & 0.7392   & 0.8999   & 0.7962   & 0.8608   & 0.6602   & 0.5929   & 0.3844   & 0.7737   & 0.7446   \\
     &        GMM\cite{Zivkovic2004}            & 0.8245   & 0.6330             & 0.5969   & 0.5207   & 0.7370   & 0.6621   & 0.7380   & 0.5373   & 0.4097   & 0.1522   & 0.4663   & 0.5707   \\
     &          3PDM \cite{2020_TITS_8782599}   & 0.8820   & 0.8990             & 0.7270   & 0.6860   & 0.8650   & 0.8410   & 0.8280 & 0.5350 & 0.4210 & 0.5010 & 0.7930 & 0.7253 \\
     &          HMAO \cite{2019_TIP_8543221}    & 0.8200   &  N/A               & 0.6300   & 0.7200   & 0.8600   & 0.8400   & 0.7900 &  0.6000 & 0.3600 &  N/A &  0.4600 & N/A \\
        &        B-SSSR\cite{2019_TIP_8485415}     & 0.9700   & \textbf{0.9500}    & 0.9300   & 0.7400   & 0.9300   & 0.8600   & 0.9200   & N/A      & N/A      & N/A      & 0.8700   & N/A      \\    
     &        MSCL-FL\cite{2017_TIP_8017547}    & 0.9400   & 0.9000             & 0.8600   & 0.8400   & 0.8600   & 0.8600   & 0.8800   & N/A      & N/A      & N/A      & N/A      & N/A      \\
     &        DSPSS\cite{2018_TPAMI_8017459}    & 0.9664   & 0.9057             & 0.8662   & 0.7870   & 0.9177   & 0.7328   & N/A      & N/A      & N/A      & N/A      & N/A      & N/A      \\
     &        STSHBM\cite{2018_TPAMI_7954680}   & 0.9534   & 0.9120             & 0.8503   & 0.8349   & 0.8930   & 0.8579   & N/A      & N/A      & N/A      & N/A      & N/A      & N/A      \\
\hline
        \parbox[t]{2mm}{\multirow{10}{*}{\rotatebox[origin=c]{90}{deep learning methods}}}   & BMN-BSN \cite{2019_WACV_mondejar2019end} & 0.9521 & 0.6371 & 0.6962 & 0.6369 & 0.7893 & 0.7849 & 0.8124 & 0.6426 & 0.6125 & N/A & N/A & N/A \\
     &   DeepBS\cite{2018_PR_BABAEE2018635} & 0.9580   & 0.8761   & 0.8990   & 0.6098   & 0.9304   & 0.7583   & 0.8301   & 0.6002   & 0.5835   & 0.3133   & 0.8455   & 0.7548   \\
     &        CNN-SFC\cite{2019_JEI_bgconv}     & 0.9497   & 0.9035   & 0.8035   & 0.7499   & 0.9127   & 0.8494   & 0.9084   & 0.7808   & 0.6527   & 0.7280   & 0.8288   & 0.8243   \\
     &                 CwisarDH\cite{6910014}   & 0.9145   & 0.8274   & 0.7886   & 0.5753   & 0.8581   & 0.7866   & 0.6837   & 0.6406   & 0.3735   & 0.3218   & 0.7227   & 0.6812   \\
     &        $\text{DPDL}_{40}$   \cite{2018_ICME_8486510}            & 0.9692 & 0.8692 & 0.8661 &  0.8759 & 0.9361 & 0.8379 &  0.8688 & 0.7078 &  0.6110 & 0.6087 & 0.7636 & 0.8106   \\
        & 3DCD-55\cite{2020_TIP_9263106}               &  0.9100 & 0.8500 & 0.8100 & 0.8300 & 0.8800 & 0.8500 & 0.9500 & 0.7400 & 0.8700 & N/A & 0.9200  & N/A \\
        & FgSN-55\cite{2018_PRL_LIM2018256}           & 0.9200  &   0.6900  &  0.7600 & 0.6100 & 0.8200 & 0.7300 & 0.7200 & 0.3200 & 0.8100 & N/A & 0.5700  & N/A \\
        & MSFS-55\cite{lim2020learning}               & 0.8000  &   0.5000  &  0.8900 & 0.7000 & 0.9300 & 0.8600 & 0.8000 & 0.5500 & 0.8200 & N/A & 0.6700  & N/A  \\
        & DVTN \cite{2020_TCSVT_9281081}       & \textbf{0.9811}   & 0.9329  & 0.9014     & \textbf{0.9595}    & 0.9467 & \textbf{0.9479}  & 0.8780   & 0.7818     & 0.7737     & 0.5957 & \textbf{0.9034}  & 0.8789 \\ 
\hline
        & ADNN$_{\text{c20fs}}$       & 0.9729   & 0.9110  &  0.8997    & 0.8857    & 0.9305  & 0.9091  & 0.8512   & 0.7370     & 0.6215     &  0.6310 & 0.8405  & 0.8355 \\ 
        & ADNN-IB$_{\text{c2fs}}$    & 0.9514 & 0.8923 & 0.8309 & 0.6989 & 0.8814 & 0.7565 & 0.8635 & 0.6507 & 0.5089 & 0.1252 & 0.7007 & 0.7146  \\
        & ADNN-IB$_{\text{c8fm}}$       & 0.9562 & 0.8748 & 0.8532 & 0.8742 & 0.9347 & 0.8568 & 0.8764 & 0.7983 & 0.6161 & 0.2409 & 0.7826 & 0.7877  \\
        &        ADNN-IB$_{\text{c20fs}}$            & 0.9797 & 0.9454 & \textbf{0.9411} & 0.9114 & 0.9537 & 0.9411 &  0.9038 & \textbf{0.8123} &  0.6940 & 0.7424 & 0.8806 & \textbf{0.8826}  \\
\hline
\end{tabular}
 \vspace{-15pt}
\end{table*}

\subsubsection{Efficiency}
Recently, a few excellent deep learning networks, such as MSFS-55 and FgSN-55, have achieved almost perfect results when the training frames and computational resources are not limited.
However, the utility of these methods is limited since creating ground-truth frames is very expensive in real applications.
In addition, when the ground-truth frames used for training are limited or unseen videos are included for evaluation, such methods no longer work perfectly.
For example, according to the results published by Lim et al. \cite{2018_PRL_LIM2018256},
the FgSN method attains over 98\% in Fm value on CDnet2014.
However, when FgSN is evaluated by the partition of training and testing frames proposed by Mandal et al. \cite{2020_TIP_9263106}, the Fm value of FgSN decreases.
Furthermore, once FgSN is applied to the unseen videos of LASIESTA, it only achieves 0.41 in Fm value, as shown in \reftab{tab_unseen_LAS}.
In contrast, the proposed approach has good efficiency and only needs limited ground-truth frames to generate excellent results.
As the quantitative results of ADNN-IB$_{\text{L3fs}}$, ADNN-IB$_{\text{S2fs}}$ and ADNN-IB$_{\text{C20fs}}$, show in \reftab{tab_fg_LAS}, \reftab{tab_fg} and \reftab{tab_unseen_SBMI2015}, respectively, the proposed approach achieves the best overall Fm value for all three datasets, and the ground-truth frames used for training only take less than 1\% of ground-truth frames of the corresponding datasets.
By comparison, most of the compared methods based on deep learning networks use many more ground-truth frames than the proposed approach.
For example, 3DCD-55 used 50\% of the ground-truth frames of a particular video for training and the remaining frames from the same for testing. Compared to all these state-of-the-art methods based on deep learning networks, the proposed approach achieves the highest Fm value with the fewest number of ground-truth frames for training. This demonstrates the efficiency of the proposed approach.
In addition, the results of ADNN$_\text{c20fs}$, which is the results of ADNN-IB$_{\text{C20fs}}$ without the improved Bayesian refinement model, demonstrates the contributions of the improved Bayesian refinement model.
As shown in \reftab{tab_fg}, the improved Bayesian refinement gives the proposed approach around 5\% improvement in Fm value.

\subsubsection{Universality}
Traditionally, the parameters of background subtraction algorithms are fixed for all testing videos. This is reasonable for real applications since parameters adjustment is time-consuming.
However, most of the methods based on deep learning actually trained various networks for different videos.
Even for a few networks proposed for unseen videos, such as 3DCD  or BSUV, various networks are still trained for videos from different categories or datasets. This is unfair for comparisons between unsupervised methods and methods based on deep learning networks.
In fact, the results of unsupervised methods can be easily improved by manually adjusting their threshold values for different videos.
Fortunately, due to the generality of distribution information as well as the efficiency of the proposed approach, one arithmetic distribution neural network can be trained for all videos. This demonstrates the universality of the proposed approach.
In order to demonstrate this, ADNN-IB$_{\text{c8fm}}$ and ADNN-IB$_{\text{C2fs}}$ are trained and evaluated.
In particular, during the testing of ADNN-IB$_{\text{c8fm}}$ and ADNN-IB$_{\text{C2fs}}$, the parameters of the networks are fixed for all testing videos.
From the quantitative evaluations on LASIESTA  and SBMI2015, shown in \reftab{tab_fg_LAS} and \reftab{tab_unseen_SBMI2015}, respectively, both ADNN-IB$_{\text{c8fm}}$ and ADNN-IB$_{\text{C2fs}}$ achieve good results compared to unsupervised state-of-the-art methods.
Since no frame from these two datasets are used for training and the parameters of the proposed approach are fixed for all testing videos, the comparisons between the proposed approach and unsupervised methods are completely fair.
This way, the universality of the proposed approach is demonstrated.

From our point of view, when ground-truth frames are limited to an acceptable level and the network parameters are fixed,
methods based on deep learning networks are appropriate for comparison with unsupervised methods.
This is because researchers also adjust the parameters of their methods to achieve the best results for a particular dataset.
During parameter adjustments, the ground-truth frames are also manually checked.
Thus, both unsupervised methods and methods based on deep learning networks include prior knowledge from ground-truth frames.
The difference is that deep learning networks directly extract the knowledge from ground-truth frames, while unsupervised methods capture the knowledge through parameter adjustments by researchers with the help of ground-truth frames.
Based on this observation, although ADNN-IB$_{\text{c8fm}}$ and ADNN-IB$_{\text{C2fs}}$ are trained by frames from the CDnet2014  dataset, they are suitable for comparisons with unsupervised methods, such as SuBSENSE or IUTIS-5 on the CDnet2014 dataset.
Note that the remaining frames used for testing takes over 99\% of the ground-truth frames of the entire dataset.
From the quantitative evaluation shown in \reftab{tab_fg},
ADNN-IB$_{\text{c8fm}}$ achieves better results than IUTIS-5 which is a combination of several excellent state-of-the-art methods.
Since the parameters of ADNN-IB$_{\text{c8fm}}$ are fixed for all testing videos and the amount of ground-truth labels used for training takes less than 1\% of the entire dataset, the comparison between ADNN-IB$_{\text{c8fm}}$ and IUTIS-5 should be considered to be fair.
Also, the good performance of ADNN-IB$_{\text{c8fm}}$ demonstrates the potential of the proposed approach in real applications, since a single well-trained network can be used for all testing videos even from different datasets, based on results shown in Tables \ref{tab_fg_LAS} and \ref{tab_unseen_SBMI2015}.

The proposed approach is implemented in Pytorch\cite{paszke2019pytorch}, and the source code is available on Github\footnote{\url{https://github.com/zhaochenqiu/UBgS_ADNNet}}.
Experiments are run on a GeForce GTX 1080 GPU processor with 8 GB memory. During training 60 epochs are set as the maximum, the learning rate is set to 0.0001 and the Adam method \cite{kingma2014adam} with default parameters is used for training.

\subsection{Limitation and Future Work}
\subsubsection{Failure Case Analysis}
Although the proposed approach achieves promising results,
there are still a few failure cases which are discussed in this section.
As shown in \reftab{tab_fg}, the proposed approach does not work well in the PTZ category where videos are obtained by a moving camera. This happens because the histograms of temporal pixels are no longer useful features since pixels do not maintain their positions when a camera moves.
Fortunately, this limitation can be addressed by using the histograms of optical flows rather than temporal pixels as the input of the proposed approach.
In addition, since the proposed approach only takes less 1\% of ground-truth frames for training, it is possible that these training frames are not enough to cover the illumination variation in videos, which results in the failure of the proposed approach on videos ``I\_IL\_01'' and ``I\_IL\_02'', as shown in \reftab{tab_fg_LAS}.
This problem can be handled by increasing the number of ground-truth frames for training.

\subsubsection{Complexity Analysis}
Currently, the proposed approach takes around 3s total time to process a frame with resolution $320\times 240$ on our computer.
In particular, the histogram generation takes 0.1727s, 
the computation of the arithmetic distribution layer takes 0.4292s, the computation of classification block followed by arithmetic distribution layers takes 0.078s,
and the improved Bayesian refinement takes 2.2524s.
Although 3s is too long for real-time applications, there are several ways to accelerate the proposed approach.
First, a C++ implementation of the proposed approach should take much less time, since python is currently used for implementing the arithmetic distribution layers and improved Bayesian refinement model.
In addition, the batch size of pixels used for processing is given as 1000, which means only 1000 pixels are processed simultaneously in one processing round, since our GPU card is GTX 1080 with 8GB memory and some of the memory has to be used for exception handling.
Once a machine with a better GPU card and larger memory is used for evaluation, the proposed approach can be easily accelerated by increasing the batch size.

\subsubsection{Future Work}
The arithmetic distribution neural network proposed in this paper is just a prototype.
The architecture of the network is very simple and the number of parameters is very small.
The network used for comparisons with state-of-the-art methods in this paper has only 0.1 million parameters, the training set takes less than 1\% of ground-truth frames from the entire dataset and no pre-trained networks are used.
However, the proposed approach still achieves promising results compared to state-of-the-art methods based on deep learning networks with many more parameters than the proposed ADNN.
Thus, there is still room for improvement in the performance of the proposed approach by increasing the size of the training dataset, the number of network parameters, and integrating pre-trained networks for features extraction.
However, due to the limitation of our computational resources, the results proposed in this paper are the best we can present.
In addition, as the comparisons demonstrated in \refsec{sec_comp_CNN} show, the proposed arithmetic distribution layers are better at distributions analysis compared to the convolutional layer. This demonstrates excellent potential for the proposed approach since distribution analysis has a wide range of applications beyond background subtraction. This can be another direction in our future work.

\section{Conclusion}
We proposed the Arithmetic Distribution Neural Network (ADNN) for background subtraction. Specifically, the arithmetic distribution layers, including the product and sum distribution layers, based on arithmetic distribution operations were proposed for learning the distributions of temporal pixels. 
Also, an improved Bayesian refinement model, based on neighborhood information with a GPU implementation, was proposed to improve the robustness and accuracy of the proposed approach. 
Utilizing the arithmetic distribution layers, histograms are considered as probability density functions. 
This probability information is used during the learning procedure of the proposed approach.
Compared to previous approaches based on deep learning networks,
the proposed approach has three advantages, including: a) being effective for unseen videos;
b) promising results are obtained using limited ground-truth frames;
c) one network can be trained for all testing videos even from different datasets.
Comprehensive evaluations compared to state-of-the-art methods showed the superior performance of the proposed approach,
and demonstrated its potential for use in practical applications.

\ifCLASSOPTIONcaptionsoff
  \newpage
\fi

\bibliographystyle{IEEEtran}  
\bibliography{ref}  
\begin{IEEEbiography}[{\includegraphics[width=1in,height=1.25in,clip,keepaspectratio]{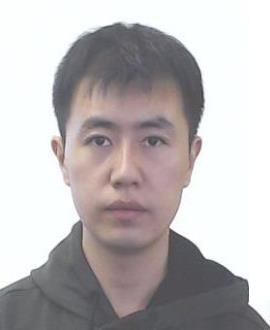}}]{Chenqiu Zhao}
received the B.S. and M.S. degrees in software engineering from Chongqing University, Chongqing, China, in 2014 and 2017, respectively.
He used to be a Research Associate with the Institute for Media Innovation from 2017 to 2018, Nanyang Technological University, Singapore .
He is currently working toward the Ph.D. degree in Multimedia Research Center, Department of Computing Science in the University of Alberta.
His current research interests include computer vision, video segmentation, pattern recognition, and deep learning.
\end{IEEEbiography}
\begin{IEEEbiography}[{\includegraphics[width=1in,height=1.25in,clip,keepaspectratio]{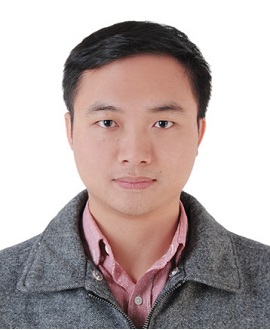}}]{Kangkang Hu}
received his Ph.D. in mechanical engineering from Carnegie Mellon University, Pittsburgh, USA, in 2016. He is currently working on UAHJIC project, in the University of Alberta. His research interests include computer vision, computer graphics, video enhancement, and deep learning.
\end{IEEEbiography}
\begin{IEEEbiography}[{\includegraphics[width=1in,height=1.25in,clip,keepaspectratio]{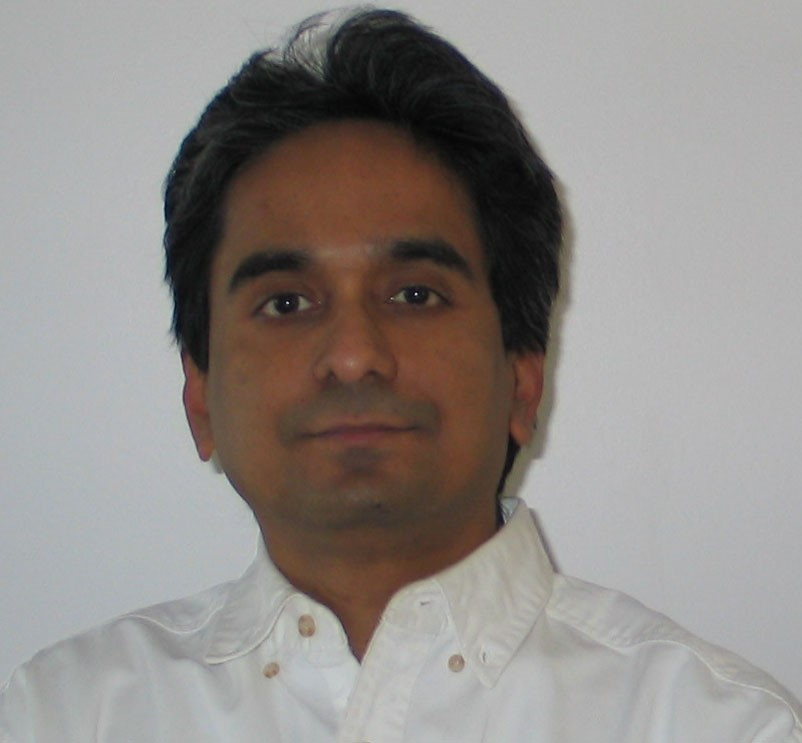}}]{Anup Basu}
received his Ph.D. in CS from the University of Maryland, College Park, USA. He originated the use of foveation for image, video, stereo and graphics communication in the early 1990s; an approach that is now widely used in industrial standards.  He pioneered the active camera calibration method emulating the way the human eyes work and showed that this method is far superior to any other camera calibration method.  He pioneered a single camera panoramic stereo, and several new approaches merging foveation and stereo with application to 3D TV visualization and better depth estimation. His current research applications include multi-dimensional Image Processing and Visualization for medical, consumer and remote sensing applications, Multimedia in Education and Games, and robust Wireless 3D Multimedia transmission. 
\end{IEEEbiography}


\end{document}